\documentclass[10pt,twocolumn,letterpaper]{article}

\usepackage[pagenumbers]{cvpr}

\usepackage{tablefootnote}
\usepackage{placeins}
\usepackage{stfloats}
\usepackage{makecell}

\definecolor{cvprblue}{rgb}{0.21,0.49,0.74}
\usepackage[pagebackref,breaklinks,colorlinks,allcolors=cvprblue]{hyperref}

\def\methodName{N4DE}
\title{Neural 4D Evolution under Large Topological Changes from 2D Images}

\author{
AmirHossein Naghi Razlighi$^{1}$ \quad Tiago Novello$^{2}$ \quad Asen Nachkov$^{1}$ \quad Thomas Probst \quad Danda Paudel$^{1}$  \\
\\
\begin{tabular}{c}
$^{1}$INSAIT, Sofia University \\
Sofia, Bulgaria
\end{tabular}
\quad
\begin{tabular}{c}
$^{2}$IMPA \\
Rio de Janeiro, Brazil
\end{tabular}
}

\begin{document}
\maketitle
\begin{abstract}

In the literature, it has been shown that the evolution of the known explicit 3D surface to the target one can be learned from 2D images using the instantaneous flow field, where the known and target 3D surfaces may largely differ in topology. We are interested in capturing 4D shapes whose topology changes largely over time. We encounter that the straightforward extension of the existing 3D-based method to the desired 4D case performs poorly.     

In this work, we address the challenges in extending 3D neural evolution to 4D under large topological changes by proposing two novel modifications. More precisely, we introduce (i) a new architecture to discretize and encode the deformation and learn the SDF and (ii) a technique to impose the temporal consistency. (iii) Also, we propose a rendering scheme for color prediction based on Gaussian splatting. Furthermore, to facilitate learning directly from 2D images, we propose a learning framework that can disentangle the geometry and appearance from RGB images. This method of disentanglement, while also useful for the 4D evolution problem that we are concentrating on, is also novel and valid for static scenes. Our extensive experiments on various data provide awesome results and, most importantly, open a new approach toward reconstructing challenging scenes with significant topological changes and deformations. Our source code and the dataset is publicly available at \href{https://github.com/insait-institute/N4DE}{https://github.com/insait-institute/N4DE}.

\end{abstract}    
\section{Introduction}
\label{sec:intro}

\begin{figure}[tb]
    \includegraphics[width=\linewidth]{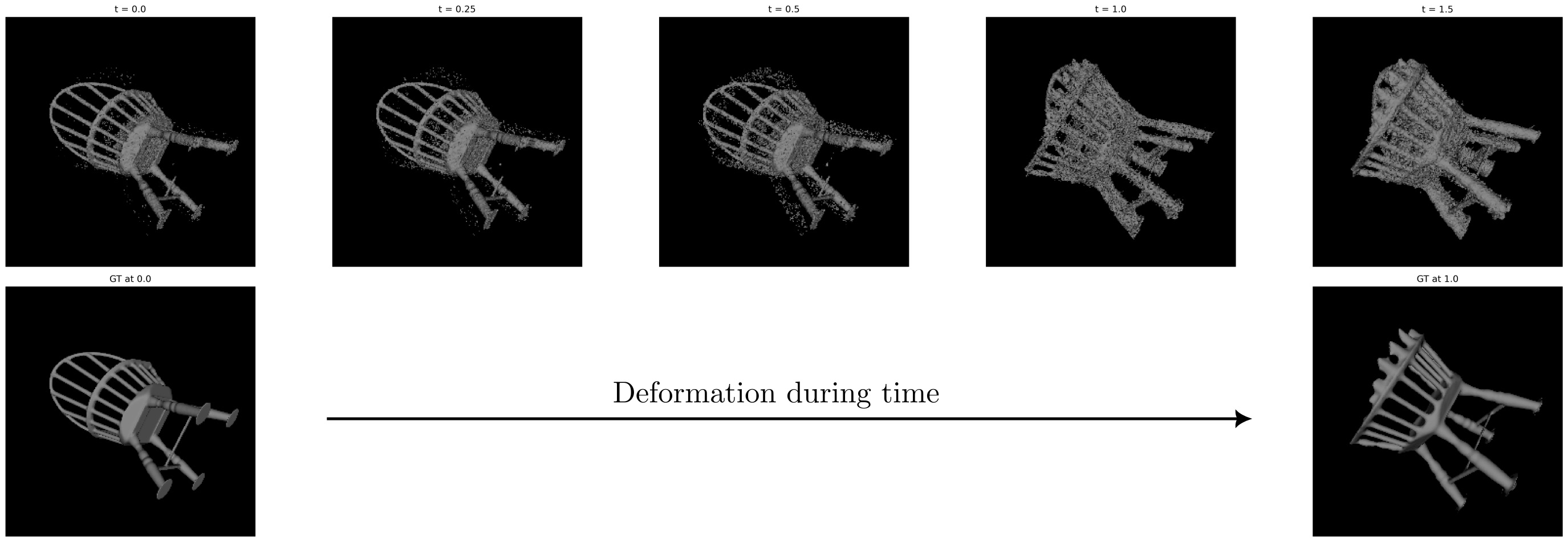}
    \caption{\textbf{Task scheme.} Our method learns the deformation animation of an object between two frames with large topological changes between them.}
    \label{fig:chair_deformation}
\end{figure}

Modeling and reconstructing the environment is an essential task for both biological and artificial systems in order to function on a higher level -- serving a variety of purposes from navigation and mapping~\cite{Taylor1992DescriptionsAD,Chiang2024MobilityVM}, understanding and interaction~\cite{Kim20133DSU,Takmaz2023OpenMask3DO3}, visualization and collaboration~\cite{Feng2023ACS}, to artistic expression~\cite{Choi2022AnimatomyAA}. 
Scanning 3D objects and scenes in particular has recently gained widespread attention due to the advent of robust algorithms for photorealistic reconstruction using handheld cameras~\cite{schoenberger2016sfm,nerf,gaussian-splatting}, and their commercial success~\footnote{See for example \href{https://www.unrealengine.com/en-US/realityscan}{Reality Scan}, \href{https://lumalabs.ai/interactive-scenes}{Luma Ai}, \href{https://ar-code.com/page/object-capture}{AR Code}}. 
Further democratization of the technology -- similar to large language models -- however is certainly limited by the static world assumption underlying those approaches~\cite{nerfies}. This leads to undesired artifacts deteriorating both the geometric accuracy and the visual quality of the reconstruction. We are therefore interested in the problem of 4D reconstruction of \textit{deformable scenes} from multiple posed RGB images. 

Extending the reconstruction algorithms to handle deforming scenes introduces additional layers of ambiguity. Those could be resolved by increasing the number of cameras and frame rate, reducing the ill-posed problem towards a series of static many-view reconstructions~\cite{Wilburn2005HighPI}. This is however neither practical nor resource efficient. Instead, an appropriate deformation model is required to enable information flow across views that may be sparse across time and space~\cite{Gao2022MonocularDV}. Existing frameworks for dynamic scene reconstruction, however, come with several caveats, such as deformation priors that are object-specific~\cite{Weng2022HumanNeRFFR,Corona2022LISALI,Paudel2024iHumanIA}, topology-restricting~\cite{hypernerf,Pumarola2020DNeRFNR,nerfies,Song2022NeRFPlayerAS}, or too weak and therefore hard to optimize~\cite{FridovichKeil2023KPlanesER,9878989}.
Moreover, depending on the underlying representation of 3D scene and its deformation, extracting the 3D surface directly is not always possible (more details can be found in Section~\ref{sec:related_works}).

We propose \textbf{\methodName}, to efficiently reconstruct the surface of a dynamic, deformable scene by supervising the model based on RGB images only. To this end, we draw inspiration from the complementary properties of implicit and explicit scene representations in three different ways:

First, we use signed distance functions (SDFs), which have proven their ability to implicitly represent complex surfaces~\cite{deep-sdf}, and have been successfully used for image-based reconstruction in conjunction with volumetric rendering~\cite{neus,Yariv2021VolumeRO,Yu2022MonoSDFEM,uni-surf,vol-sdf}. Recently, Mehta et al.~\cite{nie} showed promising results on how to perform evolution of implicit surfaces using explicit guidance. In fact, NIE demonstrates that we can incrementally update the parameters of an SDF -- implemented as a multi-layer perceptron (MLP) -- using the flow-field that is induced by an energy function that is being minimized during fitting. Performing deformations iteratively in implicit space also allows for spatially and temporally smooth transitions across  topologies~\cite{nie,Novello2022NeuralIS}. This makes it a very suitable prior in our framework.

Second, we use the well-known HashGrid encoder~\cite{instant-ngp} to discretize 3D space, thereby avoiding the slow convergence properties of fully implicit scene representations~\cite{nerf,Novello2022NeuralIS}. Given a point $(\textbf{x},t )$ in space-time, we extract its 3D latent representation of the location $\textbf{x}$ via trilinear interpolation, and feed it into an SDF module that is conditioned on the time $t$. 
We observe that this simple approach is not only able to model non-isometric deformations such as breaking a sphere, but also can evolve surfaces from one topology to another. In fact, we initialize all models in this work using the same unit sphere. Moreover, due to our choice of architecture, an explicit mesh representation can be directly obtained at any point of the continuous evolution. Importantly, this also includes interpolating and extrapolating the scene to unseen points in time.

Third, since our method already offers the surfaces explicitly, we are not bound to use expensive volumetric rendering to evaluate the photometric loss like~\cite{nerf,neus}. Instead, we propose an optional Rendering Module to place Gaussian splats directly on the mesh~\cite{sugar,Paudel2024iHumanIA}, and to use a continuous implicit parametrization for the appearance-related properties of the splats. This way, we disentangle geometry and appearance, while allowing both to continuously change over time and space. Moreover, we can enjoy the fast rendering speed and convergence of Gaussian Splatting \cite{gaussian-splatting}, while circumventing the issue of initialization and splitting or removing splats over the course of optimization.

To summarize, our contributions are as follows:
\begin{itemize}
	\item \textbf{A new architecture for reconstructing deformable objects from images:} We use a HashGrid-based \cite{instant-ngp} approach with an SDF head trained using the Neural Implicit Evolution \cite{nie} method. 
        \item \textbf{A continuous approach to Gaussian splatting.} Optionally, to predict color from target images, we estimate the splats properties on each surface point using a continuous implicit function, and  optimize it jointly with the SDF head. The approach is \textbf{initialization-free} and does not require merging or splitting splats.
	\item \textbf{Interpolation and extrapolation of deformations:} Our model is able to render the scene in unseen time steps. This fact shows that our model is actually learning the \textit{deformation} and not just overfitting on the observed frames.
    
    \item \textbf{Experiments on diverse datasets:} We evaluate our model on a diverse set of object-centric scenes with different kinds of deformations. Alongside recognized benchmark scenes, we also evaluate on  synthetic deforming animations created and rendered via Blender \cite{blender}.
\end{itemize}

\section{Related works}
\label{sec:related_works}

\begin{figure*}[ht]
    \includegraphics[width=\linewidth]{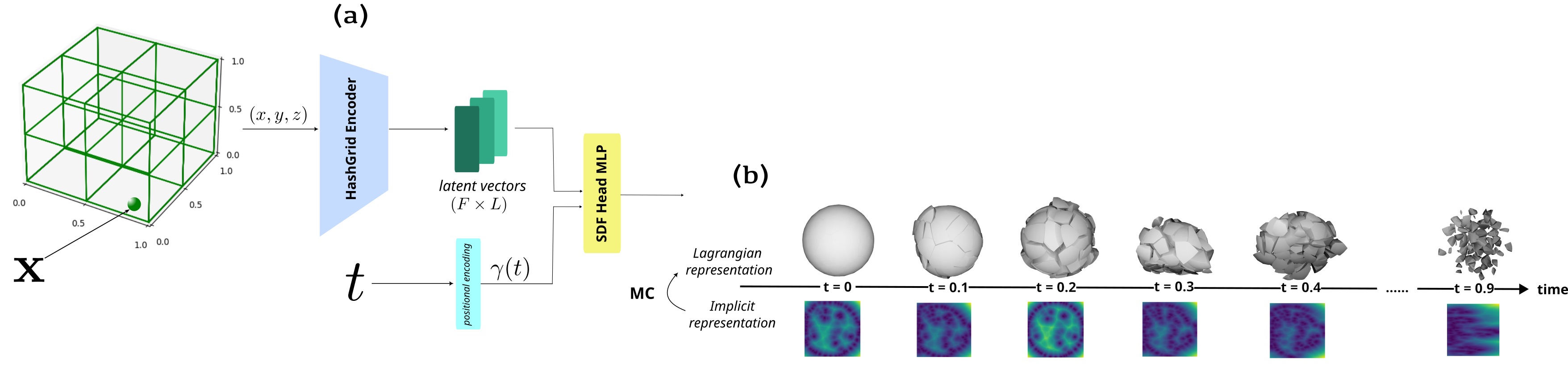}
    \caption{\textbf{Architecture of the SDF module}. Each point $\textbf{x} \in [0,1]^3$ is encoded using (a) HashGrid which is presented in Section~\ref{sec:sdf_head}. Then, the coordinate encoding (of dimension $F \times L$) are concatenated with the positional encoding of time ($\gamma(t)$) and fed into a MLP (SDF Head). (b) Signed distance value for each point is estimated, and the Lagrangian representation of mesh is extracted via Marching Cubes \cite{marching-cubes}. As our experiments show, the model can learn \textit{continuous} representation of the deformation with respect to time.}
    \label{fig:sdf_head}
\end{figure*}

Our approach is situated at the intersection of implicit and explicit  representations with regards to surface reconstruction, rendering, as well as deformation.

\textbf{Scene Representations for Differentiable Rendering}. 
In the recent literature, Neural Radiance Fields (NeRF~\cite{nerf}) and Gaussian Splatting (GS~\cite{gaussian-splatting}) have gained widespread attention. Both approach the problems of novel-view synthesis by reconstructing the 3D scene from a given set of posed RBG images. The core difference of the two methods lies in the scene representation and  rendering operation: On the one hand, NeRF is built on an implicit scene representation akin to the plenoptic function (mapping 3D coordinates and 2D viewing angle to density and colors), and requires rather expensive raymarching and sampling to evaluate the volumetric rendering integral. On the other hand, representing the scene explicitly by fitting Gaussian primitives in GS yields much faster optimization and real-time rendering speed out of the box. 
Despite the competitive visual quality however, a crucial aspect is the ability to extract accurate scene geometry. While GS often results in a lack of geometric consistency and does not offer a direct way to obtain a mesh representation of the scene without additional postprocessing or constraints~\cite{sugar,Turkulainen2024DNSplatterDA,Wolf2024SurfaceRF,Huang20242DGS}, integrating implicit surface representations like Signed Distance Function (SDF) into NeRF's volumetric rendering has proven to be a very effective strategy~\cite{neus,Yariv2021VolumeRO,Yu2022MonoSDFEM,uni-surf,vol-sdf}, directly allowing for surface extraction via MarchingCubes~\cite{marching-cubes}.
In this work, we combine implicit surface representations with  GS-based rendering to reap the benefits of both worlds. Towards this direction,~\cite{Wu2024ImplicitGS} learn continuous (implicit) splat properties on an (explicit) point cloud, yielding a compact representation that exploits similarities between neighboring splats, whereas
~\cite{Lyu20243DGSRIS,Chen2023NeuSGNI} link the SDF at the location of a Gaussian splat to its properties, leading to improved surface reconstruction. However, the problems of initializing and managing the splats and their location remain unsolved. We address those by deriving the splat location from the surface reconstructed via SDF. Moreover, our method is equipped with a powerful deformation prior to handle also non-static scenes.

\textbf{Grid Encoding}.
Learning an implicit function typically relies on encoding the input position in a higher dimension using positional embeddings~\cite{nerf} or periodic activation functions~\cite{siren}.
%hard to optimize
Although accurate, it has been shown that the convergence is rather slow and can be drastically sped up via discretization or space partitioning~\cite{Liu2023PartitionSU,instant-ngp}, typically done at multiple scales. 
Another aspect of the input space encoding is given by the dimensionality of the signal: in discrete space, the memory complexity grows exponentially with the number of dimensions. This becomes especially relevant for deformable scenes that require a time domain. To address this problem,~\cite{FridovichKeil2023KPlanesER} propose to only discretize K planes from each pair of dimensions, and ~\cite{Fang2022FastDR} only discretize 3D space, keeping the time embedding continuous.
We follow~\cite{Fang2022FastDR} for both our SDF and rendering heads due to its simplicity.

\textbf{Deformation Priors}.
Many 4D approaches do not explicitly employ a deformation prior, and
hope to compress the time-variant 3D scene without explicit guidance on the deformation~\cite{FridovichKeil2023KPlanesER,9878989}, leading to visually pleasing representations of even longer videos, but at the cost of geometric consistency, sample efficiency and convergence speed. 
On the other hand, ~\cite{hypernerf,Pumarola2020DNeRFNR,nerfies,Song2022NeRFPlayerAS} learning MLP-based coordinate warping networks jointly with a static template. For known objects such as hands and body, task-specific articulated shape models can be used~\cite{Weng2022HumanNeRFFR,Corona2022LISALI,Paudel2024iHumanIA}.
~\cite{Jang2022DTensoRFTR} use low rank approximations to factorize 4D space. 
Above approaches however are restricted to model only deformations from a single template, and cannot handle larger changes in topology. 

Inspired by Mehta et al.~\cite{nie}, we are interested in modeling a deformation as driven by a continuous evolution of the level set equation. As demonstrated by Neural Implicit Surface Evolution~\cite{Novello2022NeuralIS} (NISE), implicit neural SDFs can be time conditioned, to represent the deforming surfaces that evolve naturally without topological constraints.
Integrating NISE as a prior into our approach makes the ill-posed task of deformable reconstruction from images tractable.

\section{Method}
\label{sec:method}

\begin{figure*}[ht]
    \centering
    \includegraphics[width=\linewidth, height=0.4\textheight, keepaspectratio]{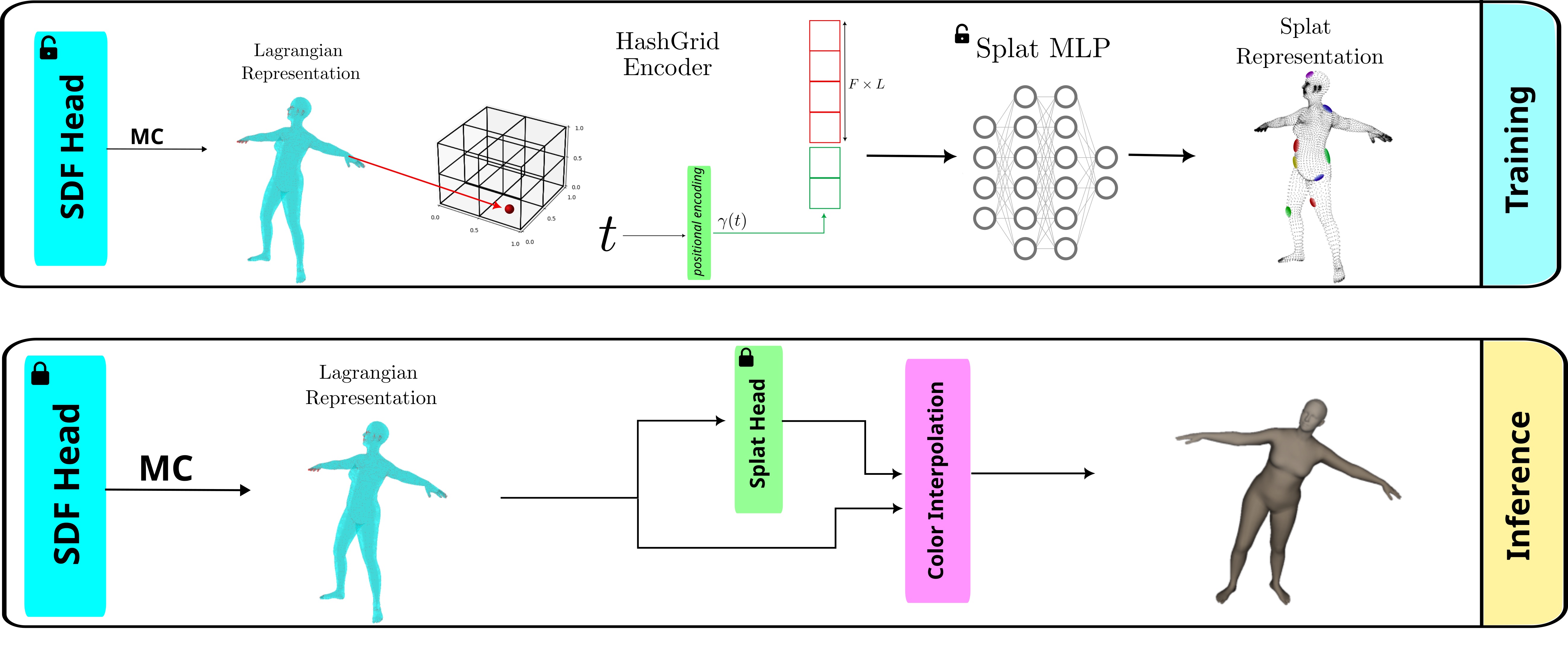}
    \caption{\textbf{Overall pipeline for training and inference with the rendering module.} In each iteration, the surface points estimated by the SDF Head are extracted via marching cubes \cite{marching-cubes}. They are then encoded via a HashGrid encoder \cite{instant-ngp} and the time embedding (via positional encoding) is concatenated to them. These features go through the rendering module to estimate the splat's appearance properties (excluding for position and scale ). At inference time, we use the final geometry and splat properties to do color interpolation and render the colored geometry.}
    \label{fig:render_head}
\end{figure*}

This section presents our model for reconstructing dynamic scenes under large topological changes from a sequence of posed RGB images. Our scene representation comprises two main components. First, we define an \textit{implicit neural representation} (INR) to model the scene geometry evolution implicitly through a dynamic \textit{signed distance function} (SDF) (Section \ref{sec:sdf_head}). Second, we introduce a rendering module, represented by a new method to perform 3DGS \cite{gaussian-splatting} via a continuous MLP estimation of the splat properties (Section~\ref{sec:rendering_head}).

\subsection{SDF Module}
\label{sec:sdf_head}
We use a HashGrid (based on Instant-NGP~\cite{instant-ngp}) to implicitly encode the geometry of the dynamic scene. A general schema of our approach towards geometry prediction based on the SDF module is available in Figure~\ref{fig:sdf_head}.
For this sake, we introduce a neural network $f_\theta:\mathbb{R}^3\times\mathbb{R}\to\mathbb{R}$ which receives both spatial coordinates $\textbf{x}$ and time $t$.
Then, for a given time step $t\in [0,1]$, $f_\theta(\textbf{x},t)$ represent the SDF value of point $\textbf{x}$ at time $t$. 
We denote the corresponding zero-level sets at each time $t$ by $S_t=\{\textbf{x}| \, f_\theta(\textbf{x},t)=0 \}$.

While HashGrid-based MLPs offer local benefits for learning 3D shapes, directly extending their domain to $\mathbb{R}^3\times \mathbb{R}$ to represent $f$ would require 4D grids, implying in costly 4D interpolations. \citet{wang2023neus2} avoids this problem by considering a 3D grid for each time step. 
However, this approach is hard to scale and suffers from high number of parameters and thus, high memory usage. 
Therefore, we take a different path based on a single 3D HashGrid encoder to overcome many of the aforementioned issues.

Precisely, we define a neural network $f_{\theta
}(\textbf{x},t)$, with $\textbf{x}$ being a point in the space and $t$ the time instance, that returns the signed distance to the surface at time $t$:
\begin{align}
\label{eq:sdf_module}
    f_\theta(\textbf{x},t) = \text{MLP}\big(E(\textbf{x}),\gamma(t)\big),
\end{align}
where $\theta$ are the MLP parameters, $E$ is the  HashGrid coordinate encoder, and $\gamma$ is a positional encoder for the time.

Since we are dealing with positional encoding of time, we scale time steps to $[0, 1]$. 
This choice will be explained further in the appendix. We extract the coordinate encoding from the HashGrid using trilinear interpolation. 
We control the speed-quality trade-off by changing the number of features $F$ and the number of levels of detail $L$. For each point $\textbf{x} \in \mathbb{R}^3$, we obtain a final encoding of size $F \times L$.

As it will become clear in the next Section~\ref{sec:evolution}, we require the difference of zero level-sets from one iteration to the next to be well behaved. Specifically, given a flow field $V^{i}(\textbf{x})$ at iteration $i$, 
\begin{align}
\label{eq:zls_bounded}
\begin{split}
   & \forall \textbf{x}\!\in\! \mathbb{R}^{3}, t \!\in\! [0,1] \,\,\exists \,\delta \!\in\!\mathbb{R}_{\geq 0}:\\ 
   &\text{CH}\big(S^{i+1}_{t}, S^{i}_{t}\big) < \delta ||V^i(\textbf{x})||_{2},  
\end{split}
\end{align}
where $\text{CH}(\cdot,\cdot)$ is the Chamfer distance, and $S^{i}_{t}$ is the zero-level set of the SDF $f_{\theta_i}(\cdot, t)$.
We show in the appendix that is indeed the case when using the coordinate encoding from the HashGrid, making the zero level-sets updates $S^{i}_t$ and $S^{i+1}_t$ move continuously between voxels.

\subsection{Implicit Surface Evolution}
\label{sec:evolution}
We build our SDF prediction head on top of the approach of Mehta et al. \cite{nie}, using the parametrized and time-conditioned SDF module, denoted as $f_\theta(\textbf{x},t)$ (see Eq.~\ref{eq:sdf_module}). To perform an iteration of the evolution, we first use marching cubes~\cite{marching-cubes} on the implicit SDF to extract a Lagrangian representation. With slight abuse of notation, we denote it as $S_t$. Then, we minimize an energy function $\varepsilon$ defined in this explicit representation. This energy induces a flow field $V(\textbf{x})$ to deform our Lagrangian as integration of the following partial differential equation (PDE):
\begin{equation}
    \label{eq:7}
    \frac{d\textbf{x}}{dt} = - \frac{\partial \varepsilon}{\partial \textbf{x}} \xrightarrow{} V(\textbf{x})
\end{equation}
In particular, we define the $\varepsilon$ to be sum of \textit{Multi-scale photometric loss} and \textit{Laplacian regularization}.
Now, to update the parameters $\theta$ of $f_{\theta}$, we need $\frac{\partial f_{\theta}}{\partial t}$. We can calculate it as:
\begin{equation}
    \frac{\partial f_{\theta}}{\partial t} = - \nabla_{\textbf{x}} f_{\theta} \hspace{0.25em} \cdot \hspace{0.25em} V
\end{equation}
Then, we compute a non-parametric \textit{next-best level-set estimate} using:
\begin{equation}
    s^{i+1} = f_\theta^i(\textbf{x}, t) - \Delta t \hspace{0.25em} \nabla_{\textbf{x}} f^{i}_\theta(\textbf{x}, t) \hspace{0.25em} \cdot V^i(\textbf{x}),
\end{equation}
which is defined on all finite vertex locations $\textbf{x}$ of the current Laplacian surface $S_t$. $\Delta t$ is a hyper parameter dependent on dynamics of the flow field. $f_{\theta}^i$ shows the SDF Module parameterized by $\theta$ in iteration $i$. Now, the loss between current estimate of SDF with next-best zero-level set can be minimized to update the model parameters as,
\begin{equation}
    L_{\text{evo}} = \frac{1}{|S_t|} \sum_{\textbf{x}_j \in S_t} ||s^{i+1}(\textbf{x}_j) - f^i_\theta(\textbf{x}_j, t)||^2.
\end{equation}
We use this loss together with additional regularizations that will be explained in Section~\ref{sec:regularizations}. An overview of the whole pipeline to estimate the geometry is shown on Fig. \ref{fig:sdf_head}.

The MLP in the SDF module is composed of $4$ hidden layers with hidden size of $128$ neurons. We use hyperbolic tangent as the activation function due to its smoothness property suitable for SDF estimation. More details about model implementation and the choice of architecture are provided in the supplementary materials.
\begin{figure}[thb]
    \centering
    \includegraphics[width=0.8\linewidth]{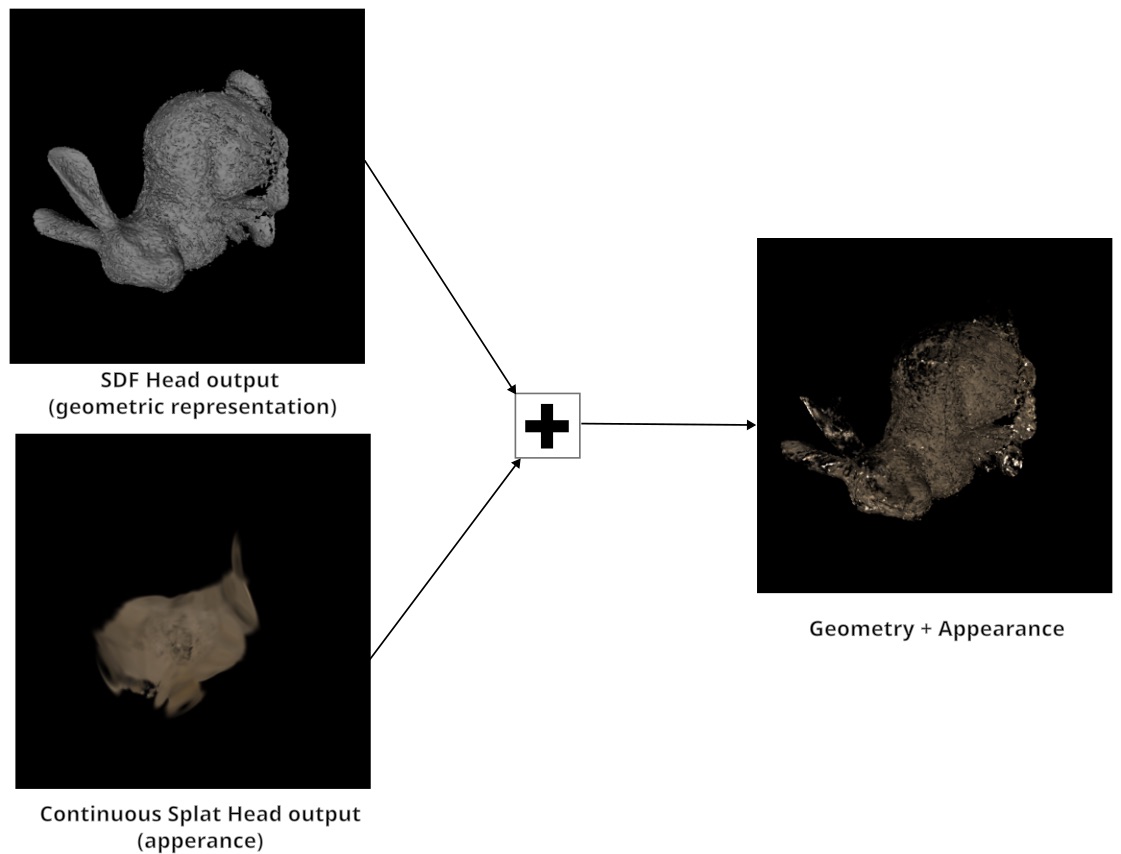}
    \caption{\textbf{Overview of the proposed pipeline, applied on a Stanford bunny.} A geometric/appearance representation is extracted using the method presented in Sec~\ref{sec:sdf_head}/Sec~\ref{sec:rendering_head}. The final colored mesh is given by the combination of these two representations.}
    \label{fig:overview_color}
\end{figure}

\subsection{Regularizations}
\label{sec:regularizations}
One of the most important aspects of our model is the shared weights throughout the training process for each of the specific time steps. This means the evolution of  e.g. $t=0.2$ affects the evolution of $t=0.1$ and vice versa. We will explain why this has a good effect in the training process for animations in which the same object is deforming. First, we use a \textit{Laplacian regularization} among with our photometric loss in the first step of our pipeline as $\varepsilon$. This is the energy function which we try to minimize and by that, we calculate the flow field to compute \textit{next-best non parametric level-set}. So, the $\varepsilon$ in \ref{eq:7} is:
\begin{align}
\begin{split}
\varepsilon &= L_{\text{photometric}} + \lambda_{\text{s}}  L_{\text{ssim}} + \lambda_{\text{l}}  L_{\text{laplacian}} \\
\end{split}
\end{align}

where $L_{\text{laplacian}}$  captures the smoothness of the surface and how evenly the vertices are distributed. We set multiplier $\lambda_{\text{l}} = 0.0002$ in the beginning and after specific number of epochs (nearly $500$ epochs), we decrease it. $L_{\text{ssim}}$ represents the SSIM loss between the estimated RGB image and the ground truth. This loss is essential for when we are doing Colored Prediction, since it introduces the structural difference between our estimate and the ground truth. In most of the experiments, we choose $\lambda_{\text{s}}$ to be $0.01$.

\begin{figure}
    \includegraphics[width=\linewidth]{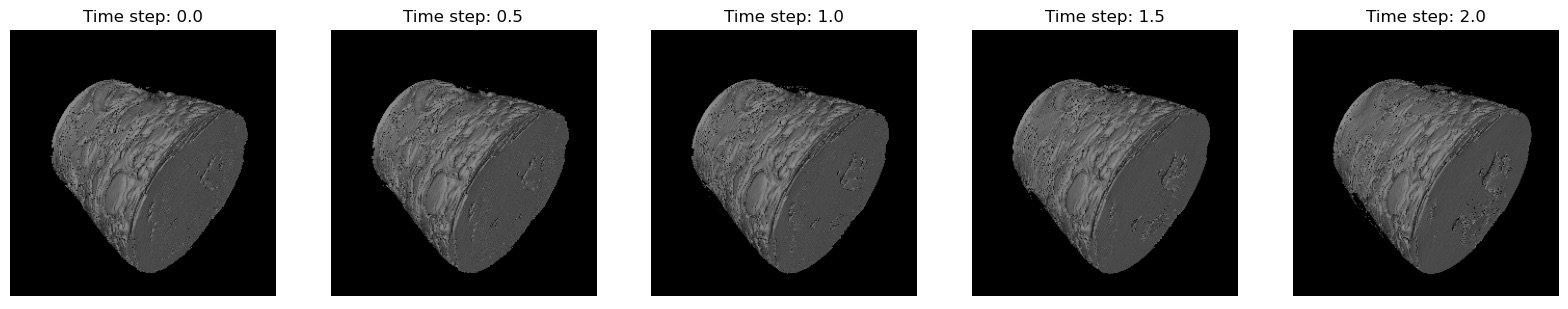}
    \caption{\textbf{Estimated meshes at different timesteps for the Static Bracelet scene.} The scene is only supervised at $t=0$ and by the effect of the $\frac{\partial S_{\theta}}{\partial t}$ regularizer, it learns to be constant along time.}
    \label{fig:bracelet_static}
\end{figure}

After obtaining the flow field, we can compute the \textit{next-best estimate of level-set} as $s(x)$. So, we use this to update our SDF module ($f_{\theta})$ parameters. The L2 loss between our current SDF estimate and $s(x)$ is called $L_{\text{evo}}$. We also add eikonal regularizaion \cite{eikonal} to find a better optimum for our SDF model. Another regularization which is used is \textit{time consistency}. Since most deformation happen over larger timescales, we aim to regularize the change within a small window of frames. For this sake, we experimented with sampling of points in time and minimizing the difference between SDF predictions of these 2 consecutive timesteps. However, we found this method to be prohibitively expensive to compute during each epoch. Instead, we opt for penalizing the deriviative of SDF w.r.t. to time. The complete loss can we written as,
\begin{align}
\label{eq:sdf_complete_loss}
\begin{split}
L = L_{\text{evo}} + \lambda_{\text{eikonal}} {\big(\lVert\nabla_{\textbf{x}} f_{\theta}\rVert - 1 \big)}^2 + \lambda_{\text{t}} \left|\frac{\partial f_{\theta}}{\partial t}\right|,
\end{split}
\end{align}
with a scheduled multiplier $\lambda_{\text{t}}$. It starts from an initial value (e.g. $0.05$) and it is damped exponentially through each training iteration. This is to emphasize time consistency between consecutive frames in the beginning, whereas later we can reduce it to capture the differences between frames in a more detailed way.

\begin{figure}[h!]
    \includegraphics[width=\linewidth]{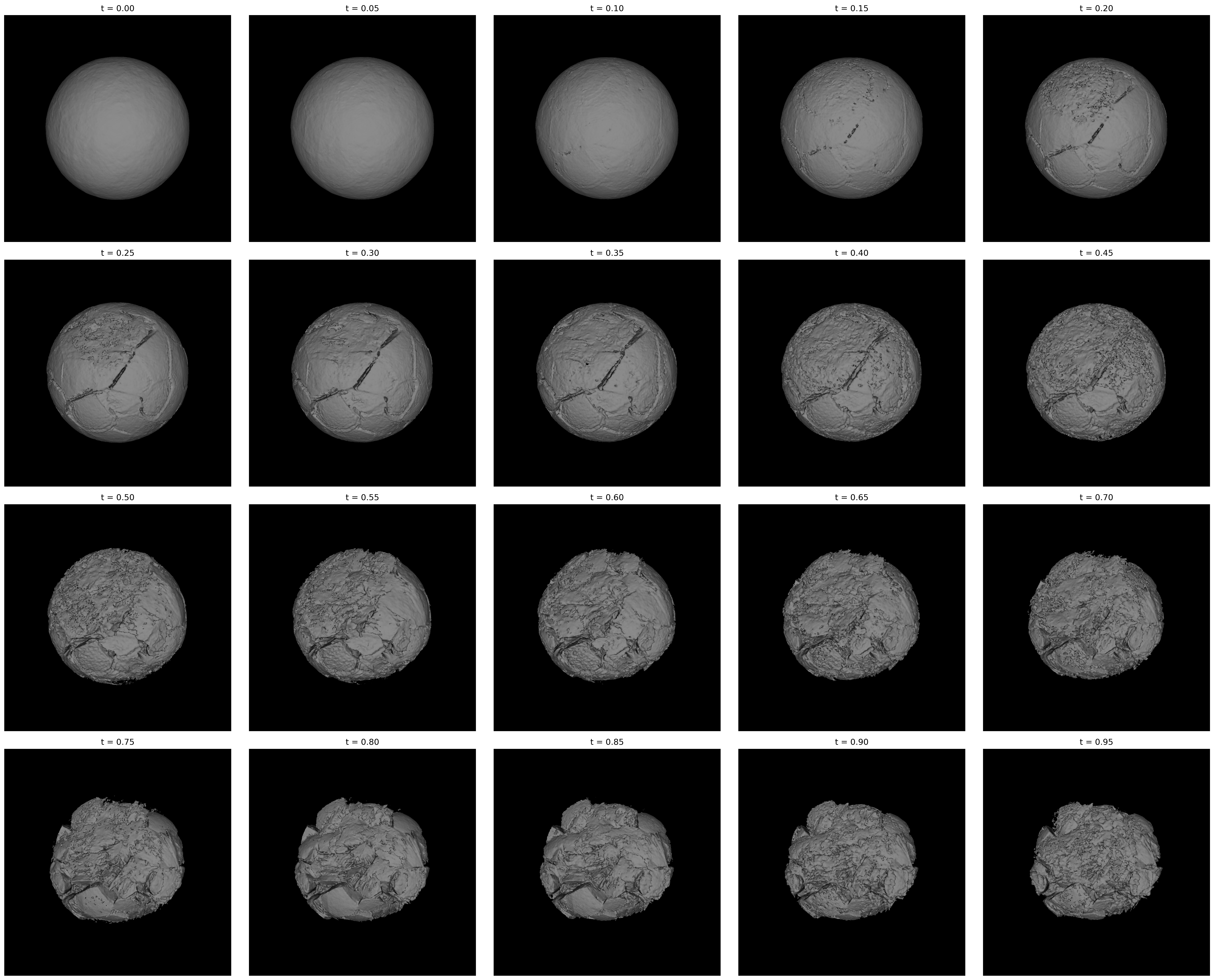}
    \caption{\textbf{Estimated meshes at different timesteps for the Breaking Sphere (Longer) scene.} The \textit{SDF Module} is trained on $5$ frames ($t=0, 0.25, 0.5, 0.75, 1$) but the total deformation animation is learned and morphing is done in unseen time-steps.}
    \label{fig:breaking-sphere}
\end{figure}

\subsection{Rendering Module}
\label{sec:rendering_head}
Our suggested pipeline is able to disentangle the geometry and appearance entirely and output two different representations for them: The SDF representation (for geometry) and the splat representation (for appearance). For the rendering module, we use a new approach toward Gaussian Splatting \cite{gaussian-splatting}. We represent the splats not explicitly, but implicitly via an MLP. In each epoch, we place the splats on the surface points of the estimated mesh (by the SDF module).
Since we are taking a \textit{continuous approach} towards Gaussian Splatting \cite{gaussian-splatting}, we need to capture the evolution that is happening in each frame for the mesh (in geometry). It is important to note that we are extracting the mesh in each epoch all over again (as explained in \ref{sec:sdf_head}). So, each vertex present in one epoch can be moved in any direction or even destroyed and replaced by another set of vertices. Our approach can cover this task and learn the splat properties implicitly. Another important aspect of our method is that the model does not need any careful initialization of the splat centroids. Actually, we initialize the splats to be on a sphere with unit radius, no matter what the target scene is, and they still converge via our method.

For this sake, we first extract the surface vertices from the SDF module. We suppose that we are placing one splat at each of these vertices. 
In the following, we omit the time parameter $t$ for clarity.
Given a SDF function $f^i_{\theta}$ trained at iteration $i$ and $\mathcal{M}$ represents the marching cubes process \cite{marching-cubes} yielding the centroids of splats as,

\begin{equation}
    V_{i} = \mathcal{M}(f^i_\theta).
\end{equation}

But since the vertices $V_{i}$  are continuously evolving and changing, we need an approach for the MLP to be able to learn this evolution implicitly. The challenge is that we do not know the mapping between the new vertices ($V_{i+1}$) and the previous set of vertices ($V_{i}$). Here we use a second HashGrid encoder~\cite{instant-ngp} to encode the surface points in an efficient, yet smooth, continuous way. 

The encoded coordinates are fed to the MLP estimator of the rendering module as $r_{\theta}(x)$ to regress the appearance features as,

\begin{equation}
   \left[ \sigma(x), \text{SH}(x), R(x) \right]= r_{\theta}(x) \quad \forall x \in V_{i}.
\end{equation}

The output contains opacity $\sigma$, spherical harmonic coefficients $\text{SH}$, and rotation $R$. The scale of the splats is not estimated, we instead set the scale of the splats to be equal to $\frac{1}{d_{vox}}$ where $d_{vox}$ is the voxel distance in our $3D$ grid. 
The complete loss to train the rendering module is then given by,

\begin{align}
\begin{split}
    L = &||I_{\text{est}}\!-\! I_{\text{gt}} ||_1 + \lambda_{\text{s}} \text{SSIM}(I_{\text{est}}, I_{\text{gt}})\\
    + &\lambda_{\text{bg}} ||I_{\text{est}}\!-\! I_{\text{bg}}||_2^2,
\end{split}
\end{align}
with $I_{\text{est}}$ and $I_{\text{gt}}$ as the estimated RGB images and ground-truth RGB images, respectively. 
$\lambda_{s}$ is  set to $0.01$ our experiments, and 
$\lambda_{\text{bg}}$ can be used in case the object is small compared to the background, to prevent all splats to collapse to be predicted as background $I_{\text{bg}}$. The reason for fixing the scale of the splats is to avoid for them to get to large and cover multiple surface points (or interior and exterior surface points). Instead, we want to get a splat representation that is geometry-aware. So that each splat should try to cover the surface point perfectly and with correct orientation, color, and opacity. While this representation is useful in many cases (for example, representing geometry via splats on the surface), it may leave some empty space between the vertices (on each face of the mesh). 

Finally in order to render the appearance to the geometry and get colored renderings, we first obtain the spherical harmonics of the surface points by via $r_{\theta}$. Then, we evaluate the spherical harmonics $ \mathcal{S}$ to obtain the RGB color of a vertex $v$ based on the viewing direction $d$ as,
\begin{equation}
    \text{RGB}(v) = \mathcal{S}(SH(v), d), 
\end{equation}
using a SH order of $3$. Then, we interpolate the colors of vertices along each face (based on barycentric coordinates) of the mesh and rasterize the image:
\begin{align}
    c_{v_p} = \lambda_1 c_{v_1} + \lambda_2 c_{v_2} + \lambda_3 c_{v_3}
\end{align}
$c$ is the color function, $v_1$ to $v_3$ are vertices of a specific triangle face of the mesh and $v_p$ is a vertex inside this face. $\lambda_i$ are barycentric coordinates on the face created by $v_i$. By this rasterization technique based on the learned $R_{\theta}$, we have 3 representations for a dynamic deformable scene: 1. Geometry (extracted from the SDF module \ref{sec:sdf_head}) 2. Surface-aligned splats 3. Enhanced RGB-colored mesh (via color interpolation)
\section{Experiments}
\label{sec:experiments}

\begin{table*}[!t]
        \caption{Quantitative measurements of SDF Module \ref{sec:sdf_head} on different scenes (geometry-only). (All reported metrics are the \textbf{average values} calculated separately for each frame.)}
        \centering
        \resizebox{\textwidth}{!}{%
	\begin{tabular}{lccccccc}
		\toprule
		Scene     & $\downarrow$MSE & $\uparrow$PSNR & $\uparrow$SSIM & $\downarrow$LPIPS & $\downarrow$Chamfer distance & Num. Frames & Epochs\\
		\midrule
            Static Stanford Bunny (single time-step)  & $0.0036$  & $24.4196$  & $0.8240$        & $0.1458$       & $0.0057$ & $1$ & $3000$\\
		Static Bracelet (single time-step)  & $0.0060$  & $22.2141$  & $0.7123$        & $0.2450$       & $0.0021$ & $1$ & $3000$\\
		Static Voronoi sphere (multi time-steps)      & $0.0044$  & $23.5453$  & $0.9087$ &        $0.0753$ & $0.0447$ & $10$ & $2440$\\
		Static Multi-Object Bunny (single time-step)      & $0.0021$  & $26.6825$  & $0.8914$ & $0.1203$  & $0.0118$ & $1$ & $3500$\\  
            Static Screaming Face (single time-step)         & $0.0076$  & $21.1543$  & $0.8345$        & $0.1710$         & $0.0167$ & $1$ & $3000$\\
		Dynamic Breaking sphere      & $0.0027$  & $29.0797$  & $0.8673$ & $0.1395$ & $0.0181$ & $3$ & $3000$ \\
            Dynamic Breaking sphere (longer)      & $0.0033$  & $26.5121$  & $0.8491$ & $0.1659$ & $0.0241$ & $5$ & $3000$ \\
            Dynamic Chair deformation      & $0.0066$  & $22.7789$  & $0.8326$ & $0.1867$ & $0.0089$ & $2$ & $5000$\\
            Dynamic Bunny deformation     & $0.0053$  & $24.5015$  & $ 0.8630$ & $0.1416$ & $0.0127$ & $3$ & $3000$\\
            Dynamic Eagle Statue deformation      & $0.0042$  & $24.2799$  & $0.8950$        & $0.1027$          & $0.0029$ & $2$ & $2000$\\
            Dynamic SMPL \cite{smpl-dataset} scene \#1         & $0.0030$  & $25.4215$  & $0.9274$        & $0.1000$          & $0.0050$ & $3$ & $3000$\\
            Dynamic SMPL \cite{smpl-dataset} scene \#2         & $0.0037$  & $24.7243$  & $0.9187$        & $0.1037$          & $0.0071$ & $3$ & $3000$\\
            Dynamic SMPL \cite{smpl-dataset} scene \#3         & $0.0052$  & $23.0766$  & $0.9071$        & $0.1307$         & $0.0275$ & $3$ & $3000$\\
		\bottomrule
	\end{tabular}%
    }
	\label{tab:quantitative}
\end{table*}

In this section, we perform experiments to evaluate our model (\methodName) and show its capabilities. The experiments are done on \textit{Nvidia A6000 RTX} and \textit{Nvidia A100 40GB}. Our appendix provides the experiment details and a comparison with NIE~\cite{nie} showing that \methodName \space greatly improves dynamic scene reconstruction. 

\begin{figure}[thb] 
    \centering
    \includegraphics[width=0.23\textwidth]{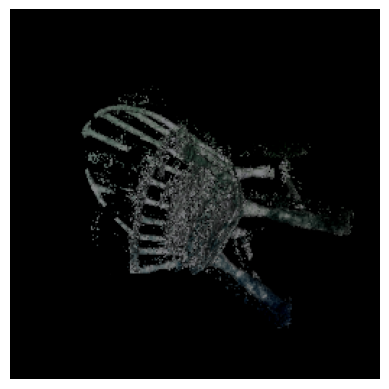}
    \includegraphics[width=0.23\textwidth]{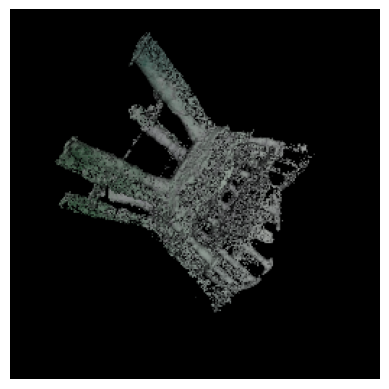} 
    \\
    \includegraphics[width=0.23\textwidth]{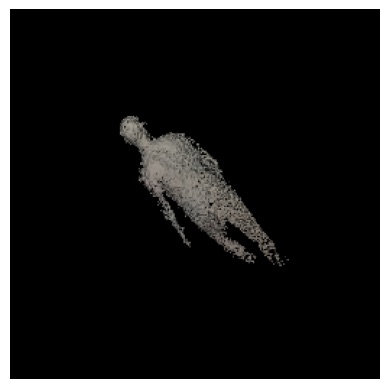}
    \includegraphics[width=0.23\textwidth]{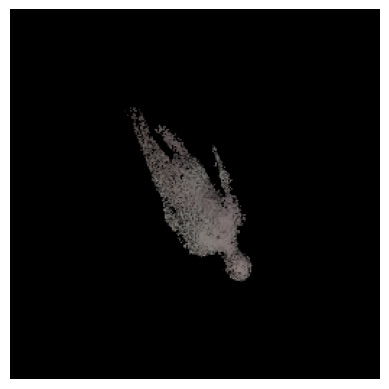} 
    \caption{\textbf{Rendered colored meshes.} We showcase \textit{Solid Colored Deforming Chair} and \textit{SMPL Scene \#2} \cite{smpl-dataset}. The splat head is trained using only $1$ random sampled image from target view.} \label{fig:colored_results}%
\end{figure}

\begin{figure}[thb]
    \centering
    \includegraphics[width=\linewidth]{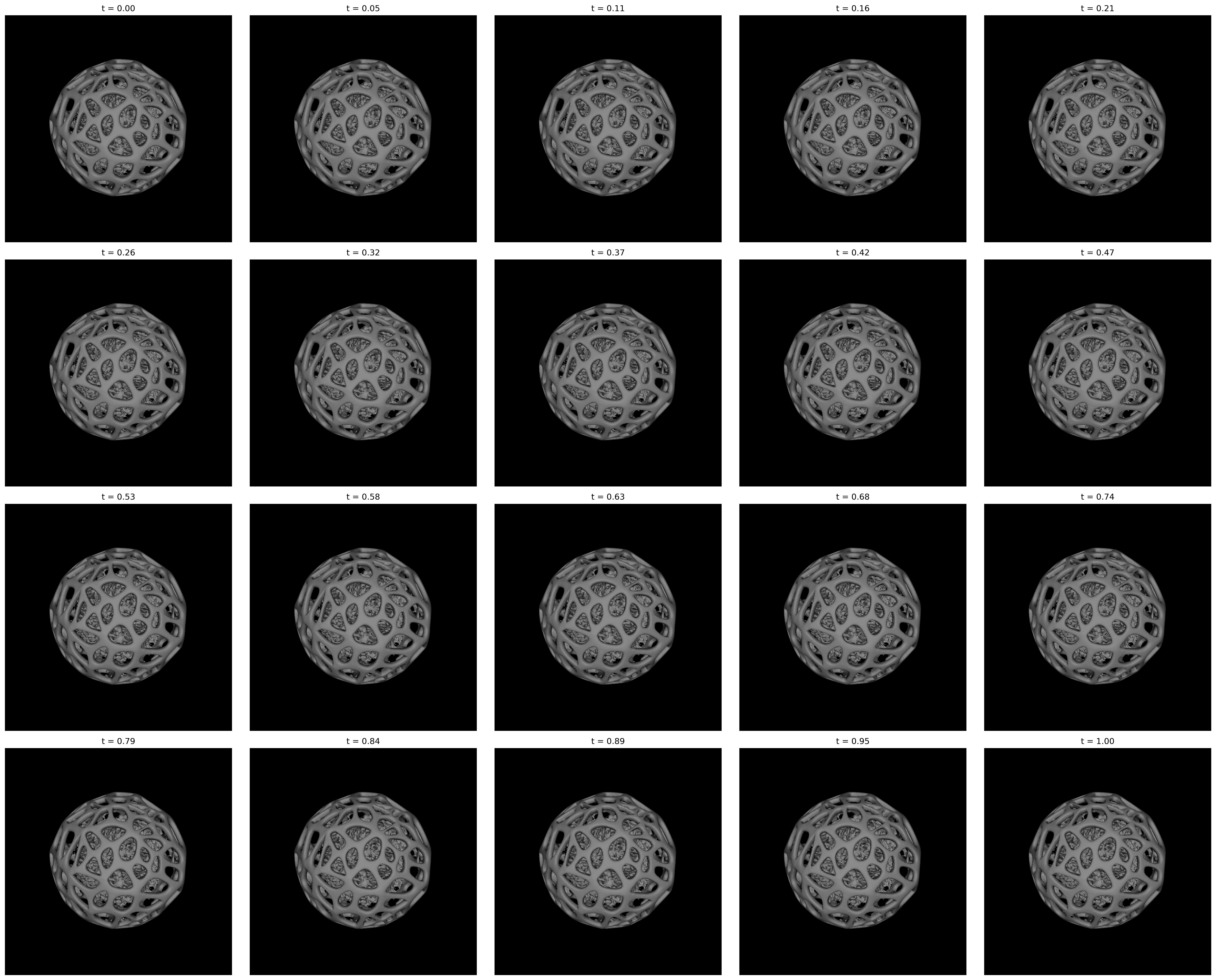}
    \caption{\textbf{Ten frame static Voronoi sphere reconstruction.} Although the model is supervised on 10 frames ($t=0, 0.1, ..., 0.9$), because of the time consistency regularization, the prediction is consistent among other times in this interval too.}
    \label{fig:voronoi}
\end{figure}

\subsection{Dynamic Deformable Scenes}
We test our method, \methodName, in several dynamic scenes, ranging from simple transformations (e.g., scaling) to complex changes (e.g., breaking). Table \ref{tab:quantitative} presents quantitative evaluations demonstrating that N4DE can effectively reconstruct challenging cases, such as the Breaking Sphere in Figure~\ref{fig:breaking-sphere} and Chair Deformation in Figure~\ref{fig:chair_deformation}. To our knowledge, our method is the first to handle such topological deformations without assumptions. For a comprehensive evaluation of efficiency and quality, we also conduct the same experiments using the NIE approach \cite{nie} (detailed in the appendix), where it frequently fails in cases where our method succeeds.

\subsection{Static Scenes}
While our model is designed to capture deforming animations over time, it is still capable of reconstructing static objects in one single frame or during different time steps. For static scenes in one single time step, we configure our \textit{HashGrid Encoder} with $F = 8$ features per-level and $L = 16$ levels, resulting in $F \times L = 128$ dimensional coordinate embedding for each point. As an example, we reconstructed the \textit{bracelet scene} with one single time step ($t=0$) and $100$ different views of resolution $256$. Also, We've reconstructed the static \textit{Voronoi sphere} scene with $n=10$ different time steps and $10$ views of each frame. The results of these experiments are available as quantitative (Tab. \ref{tab:quantitative}) and qualitative (Fig. \ref{fig:voronoi}).

\subsection{Time Consistency regularization}
For static scenes, we want the model to learn a \textit{zero deformation} animation, or simply to be static among time. We therefore put  a high weight on $\lambda_t$ of the SDF loss in Eq.~\ref{eq:sdf_complete_loss}. Note that we can use the benefits of our well-behaved continuous implicit function and calculate the derivative of the output of our model with respect to time using PyTorch's automatic differentiation \cite{pytorch}.

For the dynamic cases (with the assumption of the same object being deformed along time, with little change from two consecutive frames), we want to start from the same mesh in the beginning and after some epochs, start to capture the fine differences between frames. So, we define a damping schedule $\lambda_{t}(e) = \lambda^{0}_{t} \cdot 0.995^{e}$, which decreases exponentially over epochs $e$. $\lambda^{0}_{t}$ is the initial multiplier which in most experiments is set to $0.05$. This time consistency term even works as expected in cases of \textit{static scenes}, in which, the prediction is all consistent during every $t \in [0,1]$, even if we did not supervise the model on that specific time step. An experiment showing this statement is plotted in Fig. \ref{fig:voronoi}.

\begin{figure}[thb] \centering
    \includegraphics[width=0.23\textwidth]{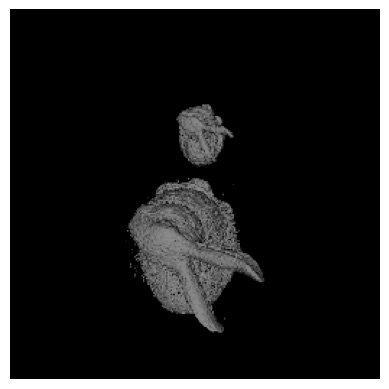}
    \includegraphics[width=0.23\textwidth,angle=-90,origin=c]{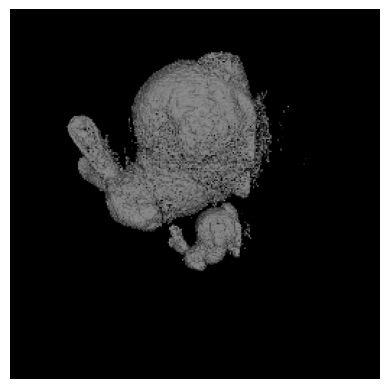}
    \caption{\textbf{Multi object reconstructions.} The \textit{Multi Object} experiment showcases that our model is also capable of evolving a simple sphere into more than $1$ object of target. The two images show the estimated geometry from $2$ different views.} \label{fig:multi-obj}
\end{figure}

\subsection{Interpolation and Extrapolation}
\label{sub:interpolation_and_extrapolation}
One important outcome of our model is that it is capable of learning the \textit{animation} and not just overfit on the supervised time-steps. It means that after training on sample time steps $t \in [i, j)$ if we infer the model on $\forall t; \hspace{0.4em} i < t < j$ we will get a meaningful mesh, with a morphing between $t_i$ and $t_j$ meshes. This fact can be seen in our dynamic experiments and static experiments (being consistent among all time-steps in a continuous manner) like Fig. \ref{fig:bracelet_static} and Fig. \ref{fig:voronoi}.

\subsection{Multi-object Reconstruction}
One of the most important aspects of our model is that it is capable of reconstructing multiple objects. 
It is worth noting that in all of the cases (including the multi-object experience) we are starting from a simple sphere and evolving into the target objects based on the 2D image loss. 
Since we are using an implicit model, the initial sphere splits into two separate bunnies with different scales, as shown in Figure~\ref{fig:multi-obj}.

\begin{table}[ht]
        \caption{Quantitative measurements of Rendering Module \ref{sec:rendering_head} on different scenes (geometry+appearance)}
        \centering
        \resizebox{\linewidth}{!}{%
	\begin{tabular}{lcccc}
		\toprule
		Scene     & $\downarrow$MSE & $\uparrow$PSNR & Num. Frames & Epochs\\
		\midrule
		Textured Stanford Bunny & $0.0091$ & $20.5140$ & $1$ & $3000$ \\
            Textured SMPL \cite{smpl-dataset} Scene \#1 & $0.0175$ & $20.1257$ & $3$ & $3000$ \\
            Textured SMPL \cite{smpl-dataset} Scene \#2 & $0.0041$ & $24.3593$ & $3$ & $3000$ \\
            Textured SMPL \cite{smpl-dataset} Scene \#3 & $0.0055$ & $22.9957$ & $3$ & $2700$ \\
            Solid Colored Deforming Chair & $0.0134$ & $19.2670$ & $2$ & $3000$ \\
		\bottomrule
	\end{tabular}%
        }
        \label{tab:render_quantitative}
\end{table}

\subsection{Implementation Details}
Our pipeline consists of a \textit{HashGrid Encoder}. The number of features per level is $F$ and number of levels in the grid encoder is $L$, resulting in a coordinate embedding of $F \times L$ dimension for each point. For high-quality results we set $F=8$ and $L=16$ resulting in a $256$ dimensional latent vector. The dimension of the look-up table $T$ is $2^{19}$ and the minimum scale of the grid is $64$ with scale of $1.5$ and $\text{linear}$ interpolation.  We can lower each of these values to make the training and inference times faster with a cost of final quality. The \textbf{SDF Head MLP} contains 4 hidden layers with $128$ neurons and $Tanh$ activation function. The input of the model is $F \times L + 64$ which is concatenation of the encoded vectors with positional encoding of time. The \textbf{Rendering Module} also consists of a HashGrid \cite{instant-ngp} exactly like the SDF Module but with a minimum grid scale of $16$ and scale of $1.3819$. The \textit{Rendering Module} has a shared backbone which processes the coordinate embeddings. This shared backbone consists of $3$ hidden layers with $ReLU$ activation and $128$ number of neurons. The output of this shared backbone is a down-sampled $64$ dimensional feature vector which is being fed to $3$ different heads: \textit{Spherical harmonics, Opacity, and Rotation}. Each of these heads is simply a linear layer with weight of dimensionality $64 \times d_{out}$. You can see the details of this Rendering module in \ref{fig:render_head}. Some of the quantitative results of rendering module's output are in the Tab. \ref{tab:render_quantitative} and some rendered samples for qualitative comparisons are presented in Fig. \ref{fig:colored_results} and Fig. \ref{fig:overview_color}.

\section{Conclusions}
\label{sec:conclusions}
We presented \methodName, a new approach towards a general way of reconstructing 4D scenes with large topological changes -like breaking and changes in the structure- via a neural evolution approach. This approach opens a new way towards modeling 4D deformations in the future. One limitation of our method is that it requires increasing the complexity of the network if we want to capture more complex deformations with high quality. This challenge is a step forward to enhance our method in the future.

\clearpage
{
    \small
    \bibliographystyle{ieeenat_fullname}
    \bibliography{main}
}

\clearpage
\setcounter{page}{1}
\maketitlesupplementary
\label{sec:appendix}

\section{Initialization scheme}
\label{sec:initialization}
For initialization, we choose a simple yet efficient initialization approach for both of our \textbf{SDF Module} (\cref{sec:sdf_head}) and \textbf{Rendering Module} (\cref{sec:rendering_head}). 

\subsection{Initializing SDF Module}
\label{sec:sdf_init}
We first initialize the SDF Module to be a sphere in all sampled time-steps between $0$ and $1$. For this sake, we sample $N$ points (in our experiments, $N = 2^{18}$) in each iteration and feed them into the SDF Module. We expect the outcome SDF values to represent a \textit{unit sphere}. For this sake, we can use the SDF equation of a unit sphere:
\begin{equation}
    s(\textbf{x}) = \sqrt{x^2 + y^2 + z^2} - 1
\end{equation}

Also, we concatenate the time $t$ as the fourth dimension with the above $N$ points. Note that $t \sim U(0, 1)$. We use a combination of a simple MSE loss function and a MAPE loss function to initialize model parameters to predict a sphere:
\begin{equation}
    \text{Loss} = || f_{\theta}(\textbf{x}, t) - s(\textbf{x}) ||_2^2 + \lambda_{mape} . \frac{f_{\theta}(\textbf{x}, t) - s(\textbf{x})}{|s(\textbf{x})|}
\end{equation}

\begin{figure}[!ht]
    \centering
    \includegraphics[width=0.15\textwidth]{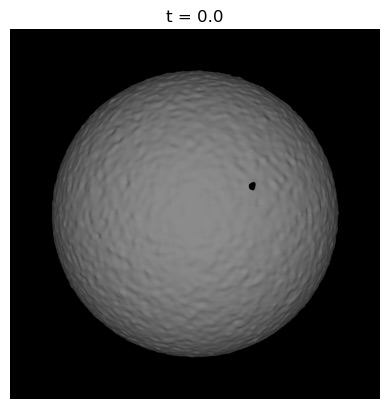}
    \includegraphics[width=0.15\textwidth]{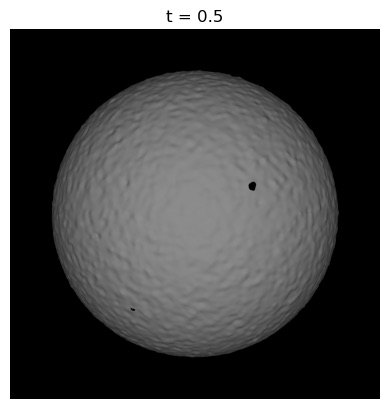}
    \includegraphics[width=0.15\textwidth]{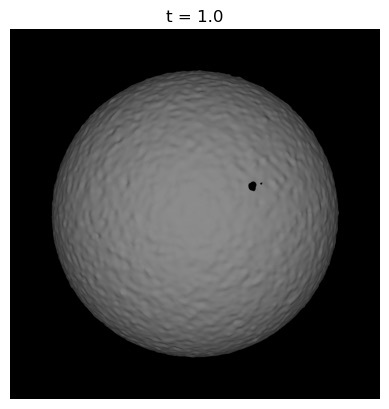}
    \caption{SDF Module (\cref{sec:sdf_head}) initialized as sphere via our initialization schema.}
    \label{fig:init_sdf}
\end{figure}

Here, $\lambda_{mape}$ is the multiplier for the MAPE loss. We set $\lambda_{mape} = 0.2$ in our experiments. You can see a sample of the model predictions ($f_{\theta}(\textbf{x}, t)$) in Fig. \ref{fig:init_sdf}.

\subsection{Initializing Rendering Module}
\label{sec:render_init}
After initializing the SDF Module to predict the sphere initially, we also fit the Rendering Module to fit the splats on the surface of these spheres in all time steps $t \in [0,1]$. Suppose $V$ is the set of vertices after marching cubes process on the zero level-set of $f_{\theta}$:
\begin{equation}
    V_t = MC(f_{\theta}(\cdot, t))
\end{equation}

\begin{figure}[!ht]
    \centering
    \includegraphics[width=0.15\textwidth]{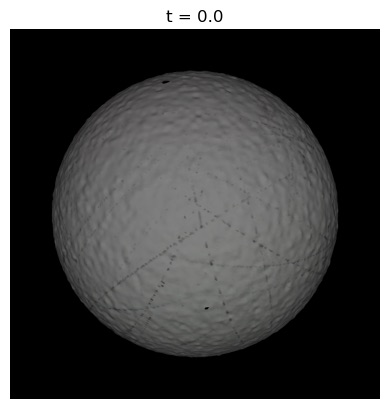}
    \includegraphics[width=0.15\textwidth]{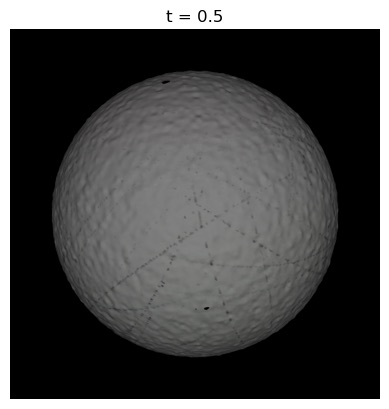}
    \includegraphics[width=0.15\textwidth]{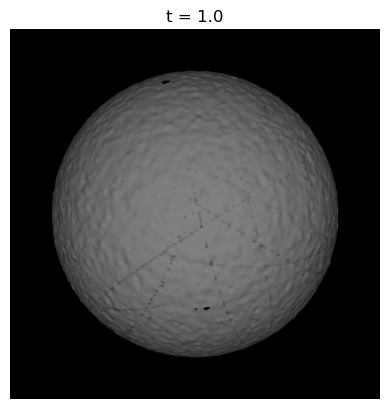}
    \caption{Rendering Module (\cref{sec:rendering_head}) outcomes after our inference approach, rendered in different time steps. The Rendering Module is initialized by placing splats on the surface of a sphere to cover the sphere completely.}
    \label{fig:init_splats}
\end{figure}

Here, $MC$ denotes the marching cubes \cite{marching-cubes} process. We position each splat on the vertices of the current Lagrangian representation ($V_t$). For scaling, we experimented with regressing the scales via the Rendering Module and also fixing them. The results have shown that pre-defining them to be $\frac{1}{d_{voxel}}$ where $d_{voxel}$ is the voxel size in our 3D sampling grid. In our experiments, based on the mesh resolution, $d_{voxel}$ can be $\frac{1}{150}$ or $\frac{1}{200}$ or $\frac{1}{256}$. Then, to fit the splats on the surface of the mesh, we define the following loss function:

\begin{equation}
    Loss = || I_{est} - I_{GT}||_1
\end{equation}

Here, $I_{est}$ represents the rasterized estimated image, and $I_{GT}$ represents the rendered (without texture) image of the sphere from the SDF Module. We know that for Gaussian Splatting \cite{gaussian-splatting} to converge, we do not need to have the correct estimate of colors necessarily. Whenever we want to infer the Rendering Module, we input the surface points (extracted from the SDF Module) and the time step ($t$) to the Rendering Module and get the predicted spherical harmonics. Then, using these spherical harmonics and the viewing direction, we render our colored mesh based on the interpolation method explained in \ref{sec:rendering_head}. The outcome of this process after initializing the Rendering Module is plotted in Fig. \ref{fig:init_splats}.

\section{Model Architecture}
\label{sec:architecture}

In this section, we will discuss the different factors that influence the outcome of our model and why we chose these architectural choices.

\subsection{The choice of HashGrid Encoder}
We chose the HashGrid encoder \cite{instant-ngp} because of two properties: 1. It fits the concept of \textit{evolution} in our SDF Module 2. It helps to learn the movements of surface points fed into the Rendering Module.

To explain further these two benefits, please note that based on the HashGrid's resolution (in each level), two points $\textbf{x}_1, \textbf{x}_2$ are encoded similarly to each other if they are close enough in the world coordinate. In mathematical form, it can be written as:
\begin{equation}
    ||E(\textbf{x}_1) - E(\textbf{x}_2)||_2 < \epsilon_1 \Longleftrightarrow ||\textbf{x}_1 - \textbf{x}_2||_2 < \epsilon_2
\end{equation}

\begin{figure}[t] \centering
    \includegraphics[width=0.46\textwidth]{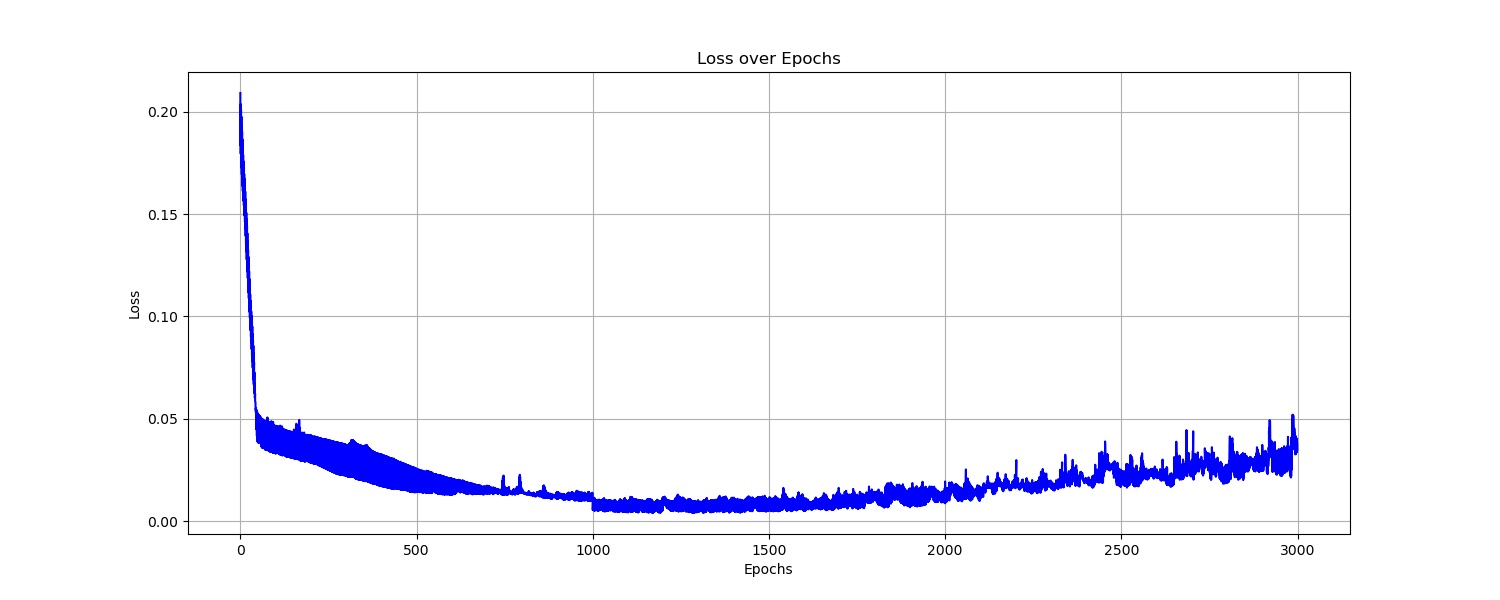}
    \caption{Loss plot (photometric loss) of NIE trained on $2$ frames of the "\textit{dynamic chair deformation}" scene. This loss plot shows the loss status of the last frame in each epoch ($t=1$). It is evident in the loss plot that the model is struggling to fit both time frames well enough, and in final epochs, as the model fits to a mesh more similar to $t=0$, the photometric loss for $t=1$ is increasing.} \label{fig:nie_loss_plot_deforming_chair}
\end{figure}

\subsection{SDF Module}
\label{sub:sdf_module}
The SDF Module consists of a HashGrid Encoder \cite{instant-ngp} with minimum resolution $N_{min}=64$, number of Hash Table entries equal to $log_2T = 19$ and $F=8$ features per HashGrid vertices and $L=16$ different levels. Please note that these properties (most notably, $F$ and $L$) can be decreased for speed. The per-level scale is set to $s=1.5$, and \textit{linear} interpolation is chosen as the interpolation method. The output of this coordinate encoder (with $F \times L$ dimension) is concatenated with the positional encoding of time $\gamma(t); \hspace{0.2em} t \in [0,1]$ with $64$ as the number of output frequencies. The concatenated vector is fed into the SDF Head MLP. The SDF Head MLP consists of $4$ hidden layers. The number of neurons per layer is set to $128$. Increasing this number to higher values (such as $256$) has increased the model's outcome in the sense of quality measures in our experiments but effectively decreases the speed of training and inference. The hyperbolic tangent ($Tanh$) is chosen here as the activation function due to its nice property of \textit{many times differentiability}, which is helpful for SDF prediction. Please refer to \cref{fig:sdf_head} for detailed explanations of the SDF module.

An explanation about why we scale $t$ to be in $[0, 1]$ is because in scenes where we use $\frac{\partial f_{\theta}}{\partial t}$, we need the time steps to be close to each other. Here, essentially, $t=0$ means the start of the animation, $t=1$ means the end of the animation, and $t \in (0, 1)$ will be some interpolation between these start and end points (which we are directly supervising in some time steps, e.g., $t=0.5$).

\begin{figure}[thb] \centering
    \includegraphics[width=0.15\textwidth]{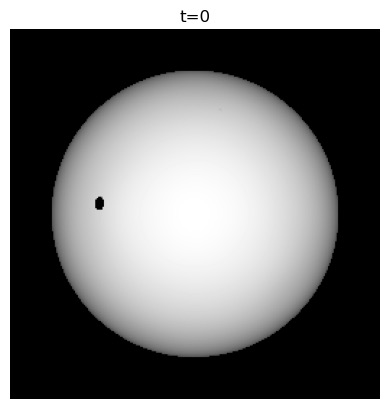}
    \includegraphics[width=0.15\textwidth]{ 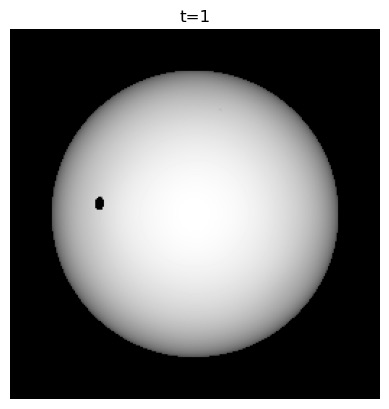} 
    \includegraphics[width=0.15\textwidth]{ 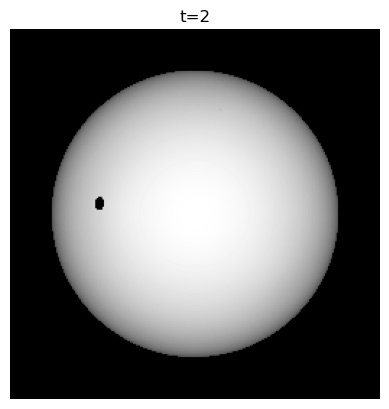}
    \caption{NIE \cite{nie} model initialized as all spheres and rendered in $3$ sample time steps.} \label{fig:nie_init}
\end{figure}

\subsection{Rendering Module}
For the Rendering Module, we also use a second separate HashGrid encoder to encode the coordinates of input points. Note that the input of this module is the estimated surface points from the SDF Module. The details of the HashGrid encoder are the same as the encoder explained in Sec. \ref{sub:sdf_module}. Number of output frequencies for encoding time ($\gamma(t)$) is $64$. The Rendering Head MLP comprises a \textbf{shared backbone} and \textbf{separate heads}. The shared backbone is an MLP consisting of $3$ hidden layer and $128$ neurons per layer. The output of this shared backbone is a $64$ dimensional feature vector. This $f \in \mathcal{R}^{64}$ feature vector is fed to $3$ separate heads to predict \textit{Spherical Harmonics, Rotation, and Opacity}. We experimented regressing \textit{Scaling} for each splat but understood that setting the scale of each splat is equal to $\frac{1}{d_{voxel}}$ where $d_{voxel}$ is the voxel size in the sampling 3D grid is more efficient.

% to backreference, use \cref{sec:xxx}

\section{Comparison with the Baseline}
\label{sec:comparison}
We compare our model (\methodName) with the baseline (NIE \cite{nie}). Since NIE is aimed to reconstruct 3D scenes via the evolution method proposed in the paper and is focused not on dynamic scenes but on static scenes, we make some changes to the code to support multi-frame scenes. We show that in many of our dynamic scenes, the outcome of our model is comparably higher quality and more aligned with the correct mesh in that time step. Also, in many cases, the NIE method fails and crashes in the training iterations. We inspected the model and understood that in some iterations, the NIE predicts all SDF values to be negative, and thus, the marching cubes step \cite{marching-cubes} fails, and the whole training pipeline crashes.

\begin{figure}[thb] \centering
    \includegraphics[width=0.23\textwidth]{ 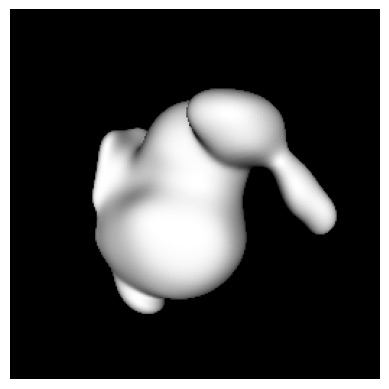}
    \includegraphics[width=0.23\textwidth]{ 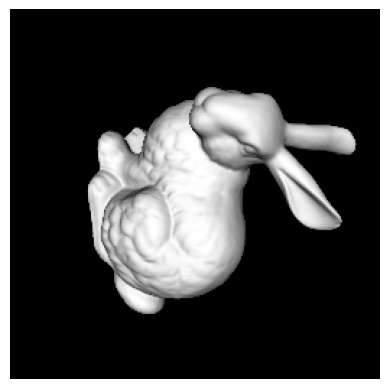} 
    \caption{NIE \cite{nie} model trained on single frame static Stanford Bunny, by giving $4D$ inputs with $t=0$ concatenated. (Left) Predicted (Right) Ground Truth.} \label{fig:nie_wtime_quality_decrease}
\end{figure}

\subsection{Extending NIE for multi-frame scenes}
We extend the NIE \cite{nie} approach to fit multi-frame scenes by inputting a 4-dimensional input instead of the previous 3-dimensional input. For this sake, we simply concatenate the time $t$ to the coordinate vector $\textbf{x}$ and then pass it to the SIREN \cite{siren} network proposed in the NIE paper. Please note that we also used the same initialization as in the main NIE paper. The initialized network's predictions in different time steps are shown in Fig. \ref{fig:nie_init}.

\subsection{Dynamic Scenes}
We compare outcomes of our model (\methodName), which are also present in \cref{tab:quantitative} with training the NIE model on the same scenes. The results are available in \ref{tab:comparison_deform}. As you can see, many of the experiments failed; thus, the table's related rows are filled with $F$. We inspected the training process and understood that NIE predicts \textit{all-negative values} for SDF, and thus, the Marching Cubes \cite{marching-cubes} process present in the training pipeline of NIE fails. This leads to \textbf{crashing the whole training pipeline}. Please note that we experimented with the initial learning rate proposed in the main paper ($lr = 0.000002$) and also with a lower learning rate. In both cases, the failing scenes still failed and crashed in the training pipeline due to the said reason.

\subsection{Static Scenes}
Adding the $4th$ dimension (time) as input to the SIREN \cite{siren} model proposed in the NIE's approach affects the output quality significantly. Aside from the dynamic scenes, even in static scenes like the Stanford bunny, we noticed that training and infering the NIE model with concatenating $t=0$ simply causes to lose a lot of fine details on the output mesh (see Fig. \ref{fig:nie_wtime_quality_decrease}).

\begin{figure}[tb] \centering
    \includegraphics[width=0.23\textwidth]{ 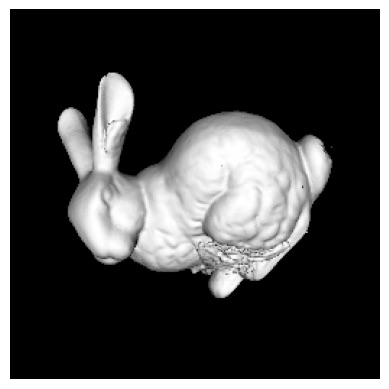}
    \includegraphics[width=0.23\textwidth]{ 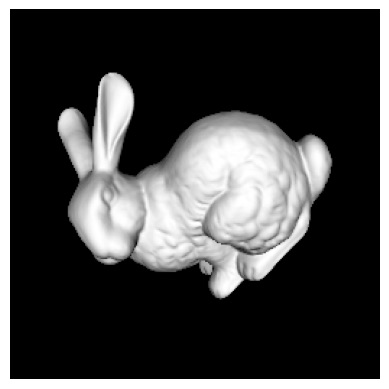} 
    \caption{NIE \cite{nie} model trained on single frame static Stanford Bunny, by giving $3D$ inputs only. (Left) Predicted (Right) Ground Truth.} \label{fig:nei_wo_time_bunny}
\end{figure}

However, doing the same scene without concatenating time $t=0$ and inputting only the $3D$ input coordinates, we get the reconstruction with good enough details and an acceptable result (See Fig. \ref{fig:nei_wo_time_bunny}). 

This essentially shows that the NIE \cite{nie} is \textbf{not capable of} handling $4D$ inputs even in static scenes, without a significant drop in the final reconstruction's quality.

\subsection{Changing NIE to accept time as 4th input}
\begin{figure}
    \begin{minipage}{0.5\textwidth}
    \centering
    \makebox[0.01\textwidth]{}
    \makebox[0.2\textwidth]{\scriptsize First time-step}
    \makebox[0.2\textwidth]{\scriptsize Second time-step}
    \\
    \raisebox{0.1\height}{\makebox[0.01\textwidth]{\rotatebox{90}{\makecell{\scriptsize Ground Truth}}}}
    \includegraphics[width=0.36\textwidth]{ 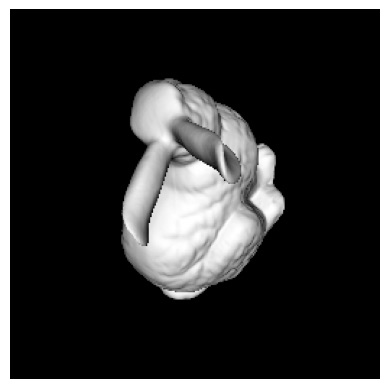}
    \includegraphics[width=0.36\textwidth]{ 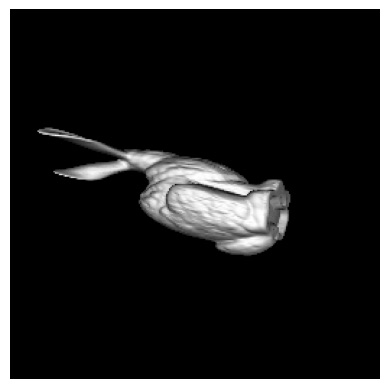}
    \\
    \raisebox{0.1\height}{\makebox[0.01\textwidth]{\rotatebox{90}{\makecell{\scriptsize NIE}}}}
    \includegraphics[width=0.36\textwidth]{ 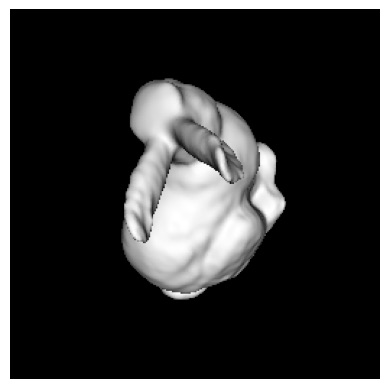}
    \includegraphics[width=0.36\textwidth]{ 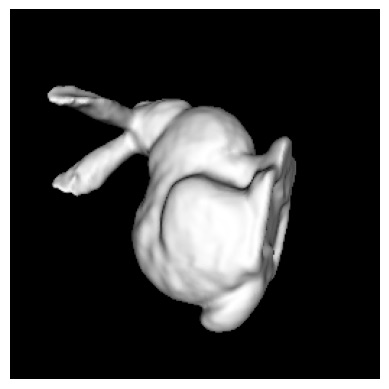}
    \\
    \raisebox{0.1\height}{\makebox[0.01\textwidth]{\rotatebox{90}{\makecell{\scriptsize \methodName(ours)}}}}
    \includegraphics[width=0.36\textwidth]{ 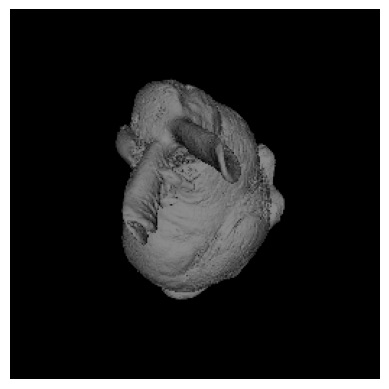}
    \includegraphics[width=0.36\textwidth]{ 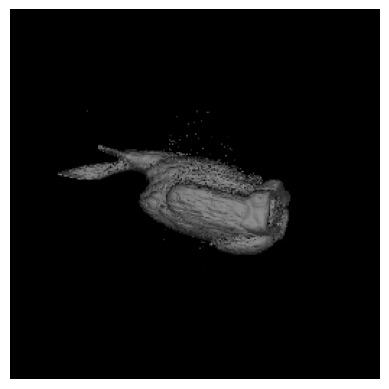}
    \\
    \caption{Visualized first two time-steps of \textit{Deformable Bunny} scene and its estimates via NIE and \methodName. As can be seen, NIE did not learn the animation properly and estimates the same mesh in different time steps, but our estimates are near the GT while distinguishing between each animation frame. Notice the Ground Truth images to observe how sudden and large the scaling deformations are over time.} 
    \label{fig:compare_deforming_bunny}
    \end{minipage}
\end{figure}

We develop our experiments more by concatenating $\gamma(t)$ instead of $t$ (where $\gamma$ stands for positional encoding). We try this approach to see if it fixes the problem of "\textit{adding a $4th$ dimension to the inputs to NIE decreases the quality in static scenes significantly and crashes the training process in dynamic scenes}". It fixes this problem, at least in the case that static scenes concatenate $t=0$ as the $4th$ dimension to the input, and the model can reconstruct the mesh with an acceptable quality similar to the original architecture. The table \ref{tab:comparison_deform} is filled with this kind of customization on NIE's \cite{nie} architecture. Even with this kind of customization on NIE's \cite{nie} architecture, most of the dynamic scenes, specifically the ones that have significant topological changes between their two consecutive frames (like SMPL \cite{smpl-dataset} scenes and the breaking sphere scene) fail due to "\textit{all negative SDF value predictions and marching cubes failure}" (refer to Tab. \ref{tab:comparison_deform}). The scenes that did not fail (like the deforming bunny, can be seen in Fig. \ref{fig:compare_deforming_bunny}), the NIE \cite{nie} model seems to overfit on one frame only or learn canonical representation between all frames instead of learning the animation itself.

\begin{figure}[b] \centering
    \includegraphics[width=0.5\textwidth]{ 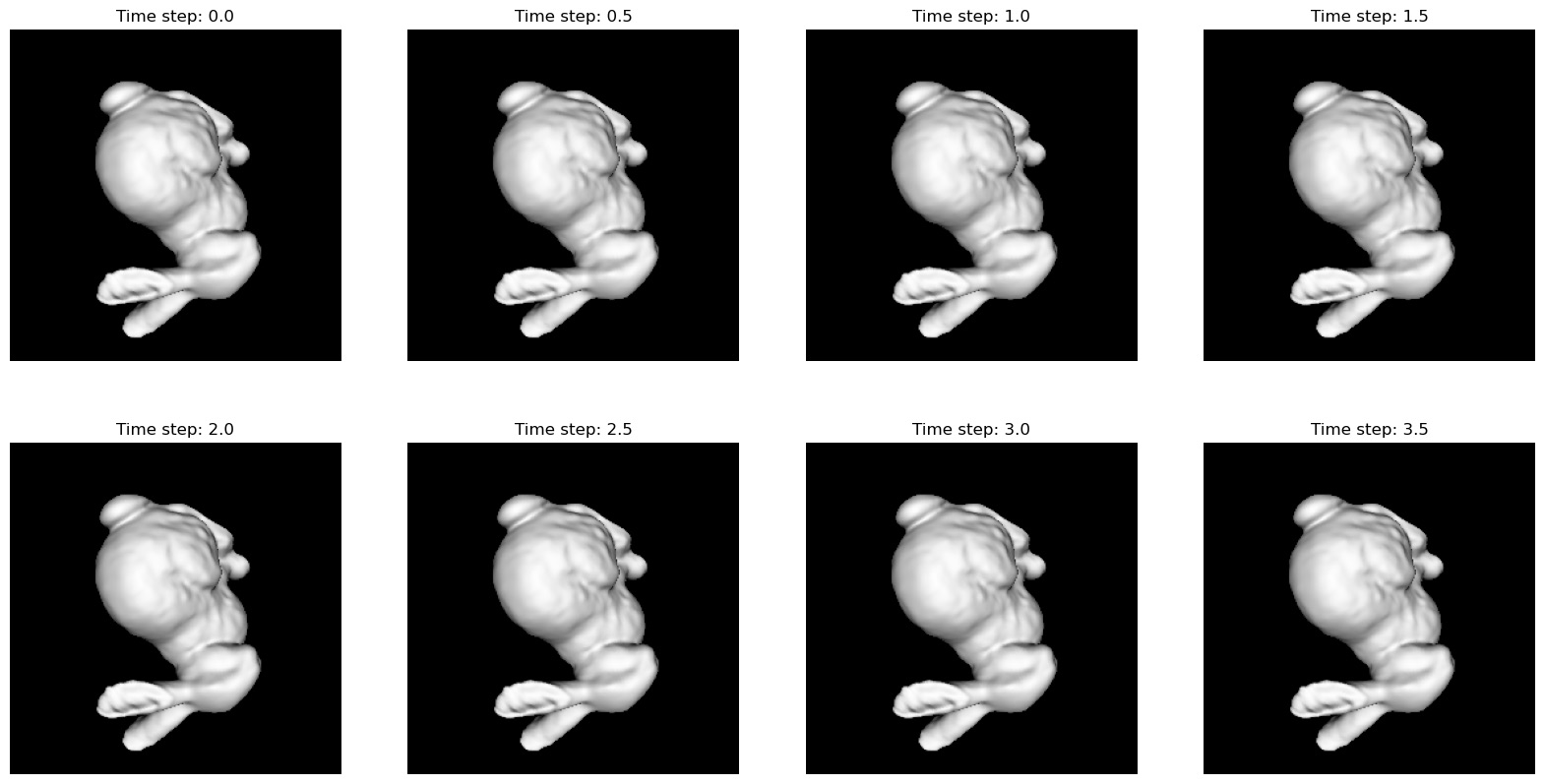}
    \caption{NIE \cite{nie} model trained on multi-frame deforming bunny scene. As you can see, the model does not learn the correct \textit{animation} conditioned on time and instead overfits to be like the first frame ($t=0$).} \label{fig:nie_deforming_bunny_grid}
\end{figure}

On the other hand, our method (\methodName) represents each time step distinctly from the other one while also being able to fill in the time gaps (to unseen frames) and interpolate between time steps. Refer to Fig. \ref{fig:compare_deforming_bunny} for a sample qualitative comparison.

\section{\methodName \space vs NIE}
In this section, we mention the main benefits of using \methodName \space instead of just customizing NIE to accept time as 4th dimension and overfit on each frame:
\begin{enumerate}
    \item It is really hard in NIE to find optimal hyper-parameters (because of using SIREN) for each scene. On the other hand, \methodName \space does not require such different configurations for each scene.
    \item NIE in the best case (concatenating $\gamma(t)$ to the $\textbf{x}$ and inputting the resulting $3 + 64$ dimensional vector to the MLP) still is incapable of representing the deformation \textbf{animation} and it just learns a mesh representation that is very similar to the ground truth in $t=0$ and looks like an average along time.
    \item NIE - even in static scenes - cannot learn a good, detailed, meaningful representation based on RGB images. It only works with high quality if we supervise it with silhouette-like gray-scale images (Images rendered with Phong Shader and without texture).
    \item \methodName \space has an obvious superiority compared to NIE regarding training time and inference time (refer to Tab. \ref{tab:comparison_deform} for some comparisons). This is because HashGrid and a much smaller MLP are used as the SDF head.
\end{enumerate}

\begin{figure}[thb] \centering
    \includegraphics[width=0.46\textwidth]{ 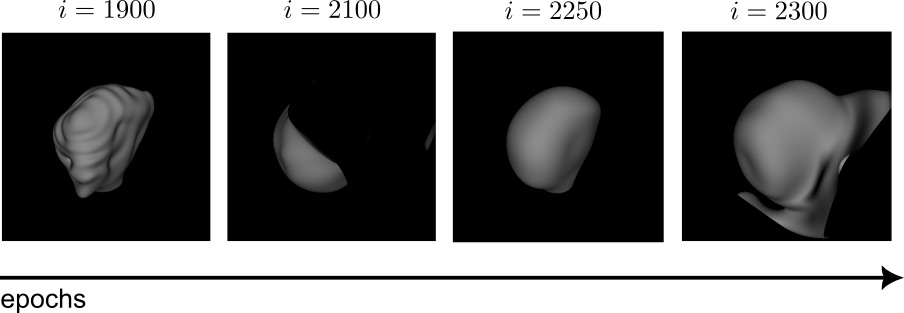}
    \caption{NIE \cite{nie} model trained on five frames of the "Deformable Breaking Sphere" scene. It failed after $2303$ epochs. The last frame ($t=5$) evolution is plotted in this figure during epochs. Compare it with our model's outcomes on the same scene in \cref{fig:breaking-sphere}} \label{fig:nie_breaking_sphere}
\end{figure}

To further investigate the outcomes of NIE, we've plotted the "\textit{Deformable Breaking Sphere}" with $5$ frames. After $2303$ epochs, this scene failed during the "all negative SDF prediction" issue. You can see the model's predictions before this failure happens in Fig. \ref{fig:nie_breaking_sphere}.

Another critical factor in our method is \textbf{training speed}. Learning animation is time-consuming, and if we want high-quality reconstructions, we may have to sacrifice speed sometimes. However, because of using the HashGrid \cite{instant-ngp} encoder, we can have a much smaller MLP as the SDF head and, thus, decrease the training time significantly. In Tab. \ref{tab:comparison_deform}, you can see the speed comparisons between our method and NIE. Please note that these time estimates are calculated based on averaging the number of seconds that took up to a specific epoch ($i$) and dividing by the number of epochs. The total sum of the seconds taken up to a specific epoch is extracted from our tensorboard \cite{tensorflow} logs for better estimates.

\begin{figure}[thb]
    \centering
    \begin{minipage}{0.5\textwidth} 
        \centering
        \includegraphics[width=0.25\textwidth]{ 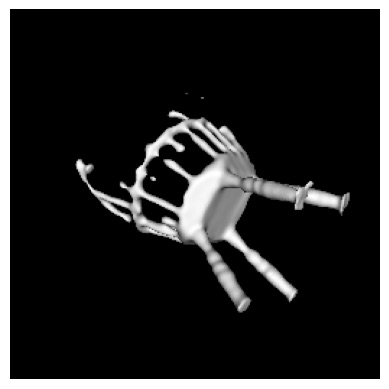}
        \includegraphics[width=0.25\textwidth]{ 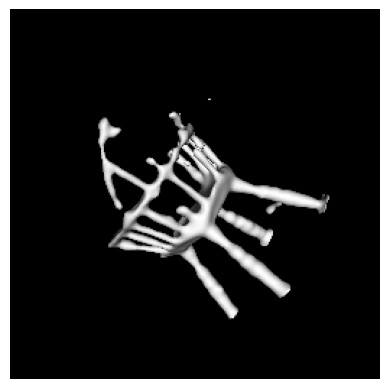}
        \\
        \includegraphics[width=0.25\textwidth]{ 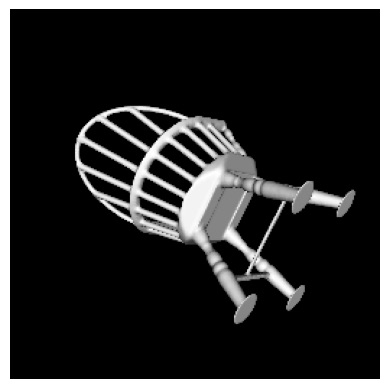}
        \includegraphics[width=0.25\textwidth]{ 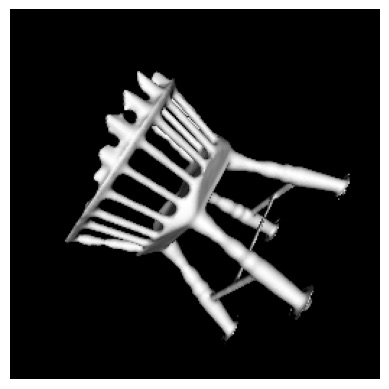}
        \caption{NIE model's reconstruction of the "\textit{Deformable chair}" scene. (Top) NIE reconstruction in $t=0$ and $t=1$. (Bottom) Ground truth in $t=0$ and $t=1$.}
        \label{fig:nie_chair_deform}
    \end{minipage}\hfill
\end{figure}

\begin{figure}[hbt] \centering
    \includegraphics[width=0.46\textwidth]{ 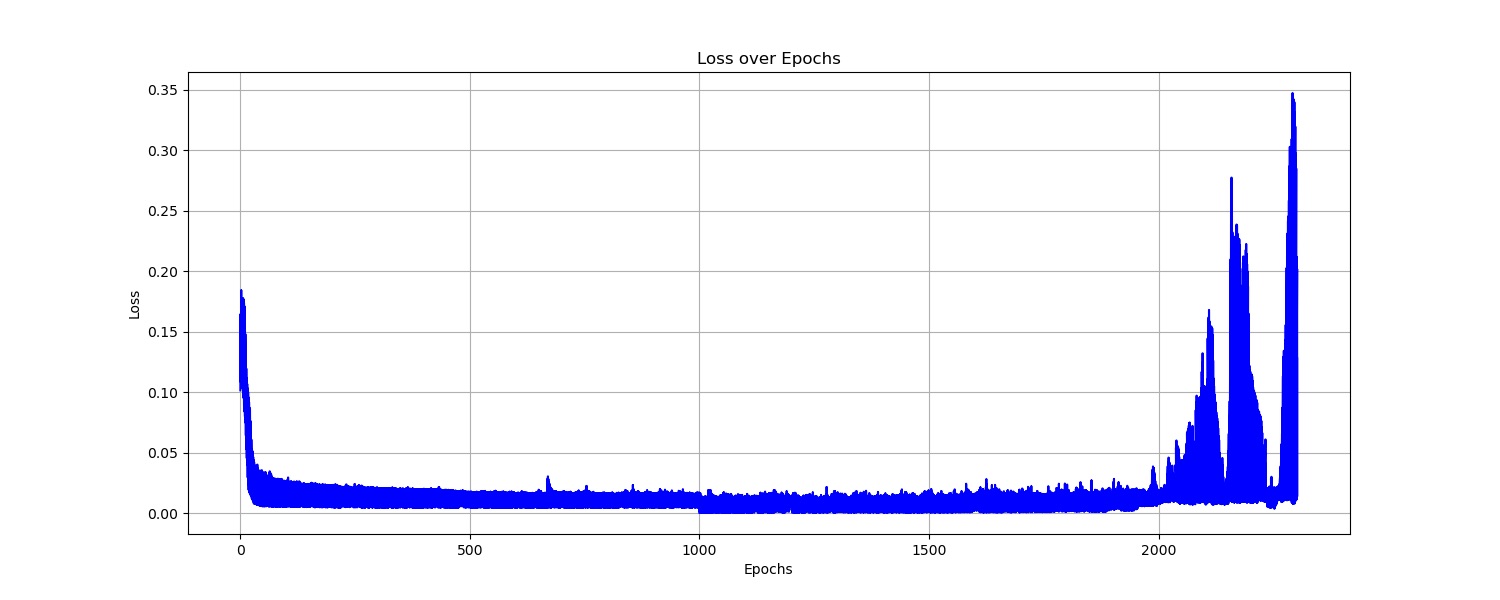}
    \caption{Loss plot (photometric loss) of NIE trained on $5$ frames of the "deformable breaking sphere" scene. This loss plot shows the loss status of the last frame in each epoch ($t=4$). This experiment failed after $2303$ epochs. This is also evident in the loss plot. As you can see, we have sudden peaks in certain epochs, the most significant ones in the last epochs (near $2300$).} \label{fig:nie_loss_plot_breaking_sphere}
\end{figure}

The most important factor our model aims at is the ability to \textbf{reconstruct deformation animations}. NIE \cite{nie} model in its pure form is incapable of doing this and fails in many animated scenes if we change the $3D$ input ($\textbf{x} = (x, y, z)$) to the $4D$ input ($(x, y, z, t)$). Even with some customizations to increase the input dimensionality and help the learning process of NIE, with inputting $(x, y, z, \gamma(t))$ where $\gamma$ stands for positional encoding, the model still cannot reconstruct and distinguish different time steps of the animation. Aside from the reconstruction output (which is so similar to the ground truth in $t=0$), the problem of "\textit{all negative SDF predictions}" and crashing the training pipeline still happens in some of the deformable scenes (refer to Tab. \ref{tab:comparison_deform} and Fig. \ref{fig:nie_breaking_sphere} and Fig. \ref{fig:nie_loss_plot_breaking_sphere}).

\begin{figure}[tb] \centering
    \includegraphics[width=0.13\textwidth]{ 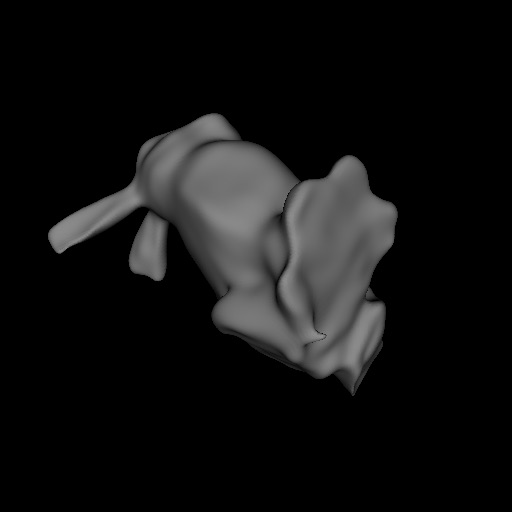}
    \includegraphics[width=0.13\textwidth]{ 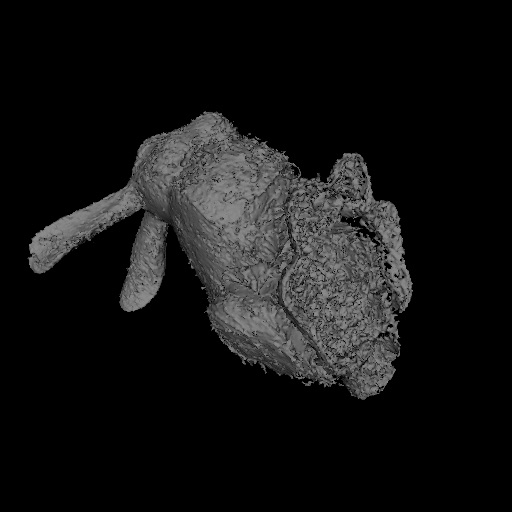}
    \includegraphics[width=0.13\textwidth]{ 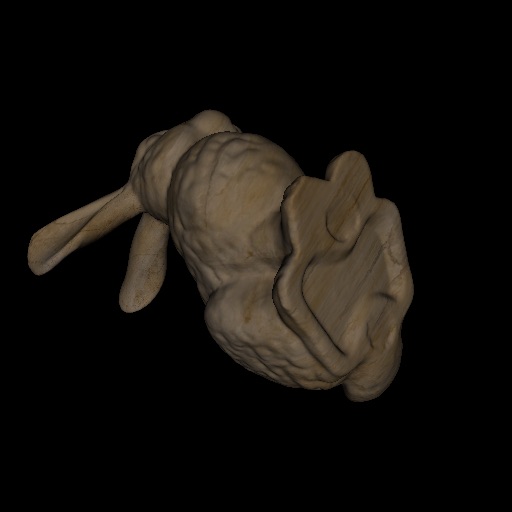}
    \caption{(Left) NIE's reconstruction when supervised via RGB textured image. (Middle) \methodName's reconstruction when supervised via RGB textured image. (Right) a sample view of the ground truth textured mesh.} \label{fig:comparison_rgb_supervision}
\end{figure}

Another difference between \methodName \space and NIE \cite{nie} is the ability to keep the reconstruction quality by changing the ground truth images. The base ground truth images used in NIE \cite{nie} are the rendered different views from the mesh by Phong Shader method and using nvdiffrast \cite{nvdiffrast} library. While this kind of supervision is useful, it is also a strong supervision method (in comparison with RGB supervision). So, we expect our model to keep its reconstruction quality on RGB (e.g., textured) ground truth images (rendered from ground truth meshes). This fact is actual for \methodName, and can be seen in \cref{fig:colored_results} and Fig. \ref{fig:comparison_rgb_supervision}. On the other hand, evaluating NIE \cite{nie} on a static scene (only in $t=0$) but with RGB supervision makes the reconstruction's quality degrade (Please refer to Fig. \ref{fig:comparison_rgb_supervision}).

\begin{table*}[thb]
    \centering
    \caption{Comparing the Baseline \cite{nie} with our approach in deformable scenes.}
    \label{tab:comparison_deform}
    \resizebox{\textwidth}{!}{
    \Huge
        \begin{tabular}{c|*{6}{c}|*{6}{c}|*{6}{c}|*{6}{c}|*{6}{c}}
        \toprule
                               & \multicolumn{6}{c}{Dynamic Bunny Deformation} 
                               & \multicolumn{6}{c}{Dynamic Breaking Sphere} 
                               & \multicolumn{6}{c}{Dynamic SMPL \cite{smpl-dataset} Scene \#1} 
                               & \multicolumn{6}{c}{Dynamic Deformable chair}  
                               & \multicolumn{6}{c}{Dynamic Deforming Statue}  \\
                               & MSE & PSNR & SSIM &  LPIPS & Chamfer dist. & Avg. time per epoch
                               & MSE & PSNR & SSIM &  LPIPS & Chamfer dist. & Avg. time per epoch
                               & MSE & PSNR & SSIM &  LPIPS & Chamfer dist. & Avg. time per epoch
                               & MSE & PSNR & SSIM &  LPIPS & Chamfer dist. & Avg. time per epoch
                               & MSE & PSNR & SSIM &  LPIPS & Chamfer dist. & Avg. time per epoch \\
        \midrule
        \methodName (ours)    & $0.0053$ & $24.5015$ & $0.8630$ & $0.1416$ & $0.0127$  & $5.8296s$ & $0.0033$ & $26.5121$ & $0.8491$ & $0.1659$ & $0.0241$  & $7.24s$ & $0.0030$ & $25.4215$ & $0.9274$ & $0.1000$ & $0.0050$  & $5.9141s$ & $0.0066$ & $22.7789$ & $0.8326$ & $0.1867$ & $0.0089$  & $2.8807s$ & $0.0042$ & $24.2799$ & $0.8950$ & $0.1027$ & $0.0029$  & $5.0204s$ \\
        \bottomrule
        NIE$^*$ \cite{nie}    & $0.0131$ & $22.2669$ & $0.8273$ & $0.1449$ & $0.0411$  & $8.4096s$ & F & F & F & F & F  & $12.03s$ & F & F & F & F & F  & $8.9504s$ & $0.0127$ & $20.1629$ & $0.8196$ & $0.1599$ & $0.0142$  & $5.47s$ & $0.0055$ & $25.4908$ & $0.9105$ & $0.0776$ & $0.0032$  & $5.4291s$ \\
    \end{tabular}
    }
    \vspace{-1em}
    \begin{flushleft}
        \footnotesize $^*\text{F}$ indicates that the training failed because of \textit{all negative} values of SDF prediction (and thus, marching cubes \cite{marching-cubes} failure). It essentially breaks the training process.
    \end{flushleft}
\end{table*}

One of the most important outcomes of our model (\methodName) is the fact that we are not just overfitting to each individual frame, but we are, in fact, learning the deformation animation by the evolution method we use from \cite{nie}. This outcome can be seen when we infer our model on time steps that the model is not supervised on (refer to \cref{fig:chair_deformation} and \cref{fig:breaking-sphere} and \cref{fig:voronoi}). This phenomenon and the effect of our time regularizer ($\frac{\partial f_{\theta}(\cdot, t)}{\partial t}$) is explained more in \cref{sub:interpolation_and_extrapolation}. The interesting fact is that in our experiments (like "\textit{Deformable Breaking Sphere}"), we found out that even if we do not use time regularization, we still get meaningful and good mesh estimates in the unseen time steps. If we want smoother, noiseless, and better quality estimates in unseen time steps, we can tune the multiplier for time regularization in the experiments. Notice that calculating and back-propagating for this regularizer adds some overhead and costs as extra time in the "Avg. training time". It's a trade-off of speed vs quality again. Even if with some more customizations and changes in the NIE's architecture (like changing the inputs more and moving them to higher dimensions, changing the order of training and randomizing the time step selection ($t$) in each epoch, etc.) in the best case it will be capable of overfitting on each frame independently. It is because nothing creates a correct relationship between each frame and the next one ($t_i$ and $t_{i+1}$) in their loss function or the architectural choice. So, eventually, it will still be incapable of learning the \textit{animation} itself (and estimating a correct mesh in the unseen time steps).

This fact can also be seen in our current experiments. If you refer to Fig. \ref{fig:nie_loss_plot_breaking_sphere} and Fig. \ref{fig:nie_loss_plot_deforming_chair}, you can see that whenever model tries to fit perfectly on one time-step ($t=i$), the other time steps' photometric loss goes up and has an increasing trend. In scenes with little changes between frames (like $2$ frame experiment on "\textit{Chair deformation}"), it will result in a reconstruction that is so similar to the first frame ($t=0$) and has some details of the $t=1$ (See Fig. \ref{fig:nie_chair_deform}). But in longer scenes like "\textit{Deformable Breaking Sphere}," the reconstruction is not exactly similar to the ground truth in $t=0$, but it looks like an average along time (See $i=1900$ in Fig. \ref{fig:nie_breaking_sphere}).

% 
% To split the supplementary pages from the main paper, you can use \href{https://support.apple.com/en-ca/guide/preview/prvw11793/mac#:~:text=Delete%20a%20page%20from%20a,or%20choose%20Edit%20%3E%20Delete).}{Preview (on macOS)}, \href{https://www.adobe.com/acrobat/how-to/delete-pages-from-pdf.html#:~:text=Choose%20%E2%80%9CTools%E2%80%9D%20%3E%20%E2%80%9COrganize,or%20pages%20from%20the%20file.}{Adobe Acrobat} (on all OSs), as well as \href{https://superuser.com/questions/517986/is-it-possible-to-delete-some-pages-of-a-pdf-document}{command line tools}.

\end{document}